\documentclass[final]{cvpr}
\usepackage{times}
\usepackage{epsfig}
\usepackage{graphicx}
\usepackage{amsmath}
\usepackage{amssymb}

\usepackage[table]{xcolor}
\usepackage{subcaption}
\usepackage{xspace}
\usepackage{multirow}
\usepackage{bbm}
\usepackage{multirow}
\usepackage{algorithm}
\usepackage[noend]{algpseudocode}
\usepackage{tabularx}
\usepackage{makecell}
\usepackage{microtype}
\usepackage{booktabs}
\usepackage{textcomp}
\usepackage{colortbl}
\usepackage{url}
\usepackage{mathalfa}
\usepackage{graphicx}
\usepackage{graphbox}
\usepackage{array}
\usepackage{enumitem}
\usepackage{animate}
\usepackage{xspace} 

\definecolor{rowblue}{RGB}{220,230,240}
\definecolor{myorchid}{RGB}{150,10,30}
\definecolor{myblue}{RGB}{10,30,250}
\definecolor{mygreen}{RGB}{10,190,10}
\definecolor{myred}{RGB}{190,20,20}

\newcommand\funop[1]{\mathop{{}#1}}

\AtBeginDocument{%
 \abovedisplayskip=6pt plus 3pt minus 5pt
 \abovedisplayshortskip=0pt plus 3pt
 \belowdisplayskip=6pt plus 3pt minus 5pt
 \belowdisplayshortskip=5pt plus 3pt minus 4pt
}

\makeatletter
\renewcommand{\paragraph}{%
  \@startsection{paragraph}{4}%
  {\z@}{0.4ex \@plus 1ex \@minus .1ex}{-1em}%
  {\normalfont\normalsize\bfseries}%
}
\makeatother

\makeatletter
\@namedef{ver@everyshi.sty}{} %
\makeatother

\newlength{\itemwidth} %
\newlength{\itemheight} %
\usepackage{tikz} %
\usetikzlibrary{calc} %
\usetikzlibrary{tikzmark} %
\usetikzlibrary{spy} %
\usetikzlibrary{shapes.misc} %
\usepackage{pgfplots} %
\usepackage{pgfplotstable} %
\pgfplotsset{compat=newest} %

\newcommand{\Lpho}{\mathcal{L}_{\text{pho}}}
\newcommand{\Lcycle}{\mathcal{L}_{\text{cyc}}}
\newcommand{\Ldata}{\mathcal{L}_{\text{data}}}
\newcommand{\Lphy}{\mathcal{L}_{\text{reg}}}
\newcommand{\Lstatic}{\mathcal{L}_{\text{static}}}
\newcommand{\Lunion}{\mathcal{L}_{\text{cb}}}

\newcommand{\Lgeo}{\mathcal{L}_{\text{geo}}}
\newcommand{\Ldisp}{\mathcal{L}_{{z}}}

\newcommand{\Wcycle}{\beta_{\text{cyc}}}
\newcommand{\Wdata}{\beta_{\text{data}}}
\newcommand{\Wreg}{\beta_{\text{reg}}}
\newcommand{\Wdisocc}{\beta_w}
\newcommand{\Wdisp}{\beta_{z}}

\newcommand{\Tjp}{T_{j}}

\newcommand{\TUnionj}{T^{\text{cb}}_{i}}
\newcommand{\blendw}{v}

\newcommand{\Pos}{\mathbf{x}} %
\newcommand{\vd}{\mathbf{d}} %
\newcommand{\vdi}{\mathbf{d}_i} %

\newcommand{\colorStatic}{\mathbf{c}} %
\newcommand{\rayStatic}{\mathbf{r}} %
\newcommand{\colorRenderStatic}{\hat{\mathbf{C}}} %
\newcommand{\colorGTStatic}{{\mathbf{C}}} %

\newcommand{\Posij}{\mathbf{x}_{i \rightarrow j}} %

\newcommand{\Posi}{\mathbf{x}_i} %

\newcommand{\pixel}{\mathbf{p}_i} %
\newcommand{\pixelDisp}{\mathbf{p}_{i \rightarrow j}} %
\newcommand{\pixelExp}{\hat{\mathbf{p}}_{i \rightarrow j}} %

\newcommand{\sfFwd}{\mathbf{f}_{i \rightarrow i + 1}} %
\newcommand{\sfBwd}{\mathbf{f}_{i \rightarrow i - 1}} %

\newcommand{\sfij}{\mathbf{f}_{i \rightarrow j}} %
\newcommand{\sfji}{\mathbf{f}_{j \rightarrow i}} %
\newcommand{\sfExpij}{\hat{\mathbf{F}}_{i \rightarrow j}} %
\newcommand{\epl}{\hat{\mathbf{X}}_{i}}

\newcommand{\depthRender}{\hat{Z}} %
\newcommand{\depthSV}{{Z}}

\newcommand{\ofij}{\mathbf{u}_{i \rightarrow j}} %

\newcommand{\sfSet}{\mathcal{F}_i} %
\newcommand{\weightSet}{\mathcal{W}_i} %

\newcommand{\wFwd}{{w}_{i \rightarrow i + 1}} %
\newcommand{\wBwd}{{w}_{i \rightarrow i - 1}} %
\newcommand{\wij}{{w}_{i \rightarrow j}} %

\newcommand{\wRender}{\hat{W}_{j \rightarrow i}} %

\newcommand{\colori}{\mathbf{c}_i}
\newcommand{\colorj}{\mathbf{{c}}_j}
\newcommand{\colorjv}{\mathbf{{c}}^{\text{cb}}_i}
\newcommand{\opacityUnion}{\sigma^{\text{cb}}}

\newcommand{\colorRenderUnionj}{\hat{\mathbf{C}}_{i}^{\text{cb}}}

\newcommand{\colorRenderj}{\hat{\mathbf{C}}_{j \rightarrow i}}
\newcommand{\colorRenderi}{\hat{\mathbf{C}}_i}
\newcommand{\colorRenderii}{\hat{\mathbf{C}}_{i \rightarrow i}}

\newcommand{\rayi}{\mathbf{r}_i}
\newcommand{\rayDisp}{\mathbf{r}_{i \rightarrow j}}

\usepackage[pagebackref=true,breaklinks=true,colorlinks,bookmarks=false]{hyperref}

\def\cvprPaperID{1264} %
\def\confYear{CVPR 2021}

\begin{document}

\title{
Neural Scene Flow Fields for Space-Time View Synthesis of Dynamic Scenes
\\[0.5cm]}

\author{Zhengqi Li${}^{1}$\qquad
Simon Niklaus${}^{2}$\qquad 
Noah Snavely${}^{1}$\qquad 
Oliver Wang${}^{2}$
\\[2mm]
${}^{1}$ Cornell Tech \qquad 
${}^{2}$ Adobe Research
}
\newcommand{\doanimated}

\twocolumn[{%
\renewcommand\twocolumn[1][]{#1}%
\maketitle
\centering
\setlength{\itemheight}{2.97cm}
\ifdefined\doanimated
\animategraphics[autoplay,loop,trim = 20 0 150 0,height=\itemheight]{20}{figures/anims/dog/}{00001}{00060} 
\animategraphics[autoplay,loop,trim = 40 0 140 0,height=\itemheight]{20}{figures/anims/bubble/}{00001}{00120} 
\animategraphics[autoplay,loop,height=\itemheight]{20}{figures/anims/womenturn/}{00001}{00120} 
\animategraphics[autoplay,loop,trim = 30 0 0 0,height=\itemheight]{20}{figures/anims/merged/}{00001}{00120} 
\else
    \includegraphics[clip,trim = 20 0 150 0,height=\itemheight]{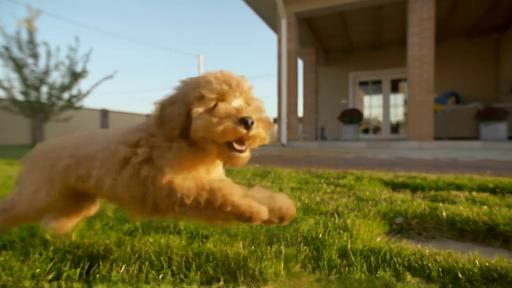} 
    \includegraphics[clip,trim = 40 0 140 0,height=\itemheight]{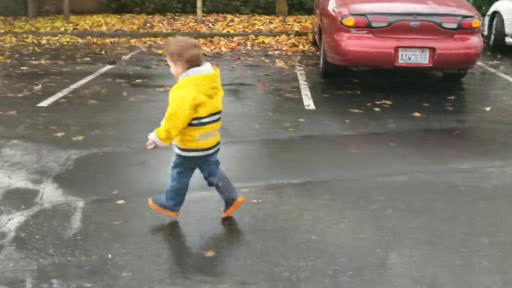} 
    \includegraphics[height=\itemheight]{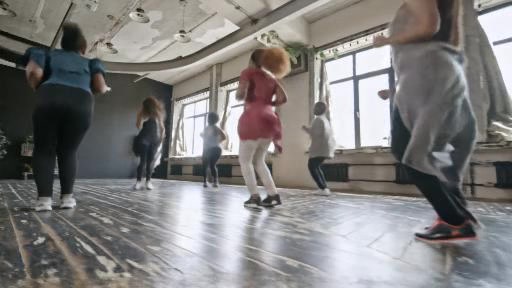} 
    \includegraphics[clip,trim = 30 0 0 0,height=\itemheight]{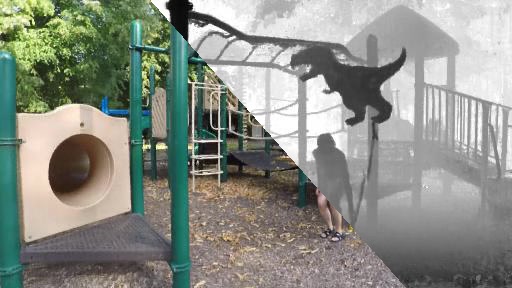} 
\fi
\vspace{-0.5cm}\small{\captionof{figure}{\label{fig:teaser} Our method can synthesize novel views in both space and time from a single monocular video of a dynamic scene. Here we show \textbf{video} results with various configurations of fixing and interpolating view and time (left), as well as a visualization of the recovered scene geometry (right). Please view with Adobe Acrobat or KDE Okular to see animations.}}
\vspace{1.2em}
}]

\begin{abstract}
We present a method to perform novel view and time synthesis of dynamic scenes, requiring only a monocular video with known camera poses as input.
To do this, we introduce Neural Scene Flow Fields, a new representation that models the dynamic scene as a time-variant continuous function of appearance, geometry, and 3D scene motion. 
Our representation is optimized through a neural network to fit the observed input views.
We show that our representation can be used for varieties of in-the-wild scenes, including thin structures, view-dependent effects, and complex degrees of motion.
We conduct a number of experiments that demonstrate our approach significantly outperforms recent monocular view synthesis methods, and show qualitative results of space-time view synthesis on a variety of real-world videos.

\end{abstract}

\section{Introduction}

The topic of novel view synthesis has recently seen impressive progress due to the use of neural networks to \emph{learn} representations that are well suited for view synthesis tasks. 
Most prior approaches in this domain make the assumption that the scene is \emph{static}, or that it is observed from multiple synchronized input views.
However, these restrictions are violated by most videos shared on the Internet today, which frequently feature scenes with diverse dynamic content (e.g., humans, animals, vehicles), recorded by a single camera.

We present a new approach for novel view and time synthesis of dynamic scenes from monocular video input with known (or derivable) camera poses.
This problem is highly ill-posed since there can be multiple scene configurations that lead to the same observed image sequences. 
In addition, using multi-view constraints for moving objects is challenging, as doing so requires knowing the dense 3D motion of all scene points (i.e., the ``scene flow''). 

In this work, we propose to represent a dynamic scene as a continuous function of both space and time, where its output consists of not only reflectance and density, but also \emph{3D scene motion}.
Similar to prior work, we parameterize this function with a deep neural network (a multi-layer perceptron, MLP), and perform rendering using volume tracing~\cite{mildenhall2020nerf}.
We optimize the weights of this MLP using a scene flow fields warping loss that enforces that our scene representation is temporally consistent with the input views. 
Crucially, as we model dense scene flow fields in 3D, our function can represent the sharp motion discontinuities that arise when projecting the scene into image space, even with simple low-level 3D smoothness priors. Further, dense scene flow fields also enable us to interpolate along changes in both space \emph{and} time. To the best our knowledge, our approach is the first to achieve novel view and time synthesis of dynamic scenes captured from a monocular camera.

As the problem is very challenging, we introduce different components that improve rendering quality over a baseline solution. Specifically, we analyze scene flow ambiguity at motion disocclusions and propose a solution to it. We also show how to use data-driven priors to avoid local minima during optimization, and describe how to effectively combine a static scene representation with a dynamic one which lets us render views with higher quality by leveraging multi-view constraints in static regions.

In summary, our key contributions include: (1) a neural representation for space-time view synthesis of dynamic scenes that we call \textit{Neural Scene Flow Fields}, that has the capacity to model 3D scene dynamics, and (2) a method for optimizing Neural Scene Flow Fields on monocular video by leveraging multiview constraints in both rigid and non-rigid regions, allowing us to  synthesize and interpolate both view and time simultaneously.

\section{Related Work}
Our approach is motivated by a large body of work in the areas of novel view synthesis, dynamic scene reconstruction, and video understanding. 

\paragraph{Novel view synthesis.}
Many methods propose first building an explicit 3D scene geometry such as point clouds or meshes, and rendering this geometry from novel views~\cite{buehler2001unstructured,chaurasia2013depth,debevec1996modeling,Hedman2017Casual3P,klose2015sampling,snavely2006phototourism}.
Light field rendering methods on the other hand, synthesize novel views by using implicit soft geometry estimates derived from densely sampled images~\cite{plenopticsampling00,gortler1996lumigraph,levoy1996light}.
Numerous other works improve the rendering quality of light fields by exploiting their special structure~\cite{Davis2012UnstructuredLF,penner2017soft,Shi2014LightFR,Vagharshakyan2015LightFR}.  
Yet another promising 3D representation is multiplane images (MPIs), that have been shown to model complex scene appearance~\cite{broxton2019lowcost,broxton2020immersive,choi2019extreme,flynn2019deepview,mildenhall2019local,srinivasan2019pushing}. 

Recently, deep learning methods have shown promising results by \emph{learning} a representation that is suited for novel view synthesis. 
Such methods have learned additional deep features that exist on top of reconstructed meshes~\cite{hedman2018deep, Riegler2020FreeVS, thies2019deferred} or dense depth maps~\cite{flynn2016deepstereo,xu2019deepviewsyn}. 
Alternately, pure voxel-based implicit scene representations have become popular due to their simplicity and CNN-friendly structure~\cite{chen2019neural,eslami2018neural,lombardi2019neural,sitzmann2019deepvoxels,sitzmann2019scene,thies2019deferred}.
Our method is based on a recent variation of these approaches to represent a scene as neural radiance field (NeRF)~\cite{mildenhall2020nerf}, which model the appearance and geometry of a scene implicitly by a continuous function, represented with an MLP. 
While the above methods have shown impressive view synthesis results, they all assume a static scene with fixed appearance over time, and hence cannot model temporal changes or dynamic scenes. 

Another class of methods synthesize novel views from a \emph{single} RGB image.
These methods typically work by predicting depth maps~\cite{li2019learning,Niklaus20193DKB}, sometimes with additional learned features~\cite{Wiles2020SynSinEV}, or a layered scene representation~\cite{shih20203d,single_view_mpi} to fill in the content in disocclusions.
While such methods, if trained on appropriate data, can be used on dynamic scenes, this is only possible on a per-frame (instantaneous) basis, and they cannot leverage repeated observations across multiple views, or be used to synthesize novel times.

\paragraph{Novel time synthesis.}
Most approaches for interpolating between video frames work in 2D image space, by directly predicting kernels that blend two images~\cite{Niklaus_CVPR_2017_adaconv, Niklaus_ICCV_2017_sepconv, Niklaus_ARXIV_2020_resepconv}, or by modeling optical flow and warping frames/features~\cite{Bao_CVPR_2019_dain, Jiang_CVPR_2018_super, Niklaus_CVPR_2018_ctxsyn, Niklaus_CVPR_2020_softsplat}. 
More recently, Lu~\etal~\cite{Lu2020LayeredNR} show re-timing effect of people by using a layered representation.
These approaches generate high-quality frame interpolation results, but operate in 2D and cannot be used to synthesize novel views in space.

\paragraph{Space-time view synthesis.} 
There are two main reasons that scenes change appearance across time.
The first is due to illumination changes; prior approaches have proposed to render novel views of single object with plausible relighting~\cite{Bi2020NeuralRF, bi2020deep, bi2020deep2}, or model time-varying appearance from internet photo collections~\cite{Li_crowd_2020, MartinBrualla2020NeRFIT, Meshry2019NeuralRI}. However, these methods operate on static scenes and treat moving objects as outliers.

Second, appearance change can happen due to 3D scene motion.
Most prior work in this domain~\cite{bansal20204d, Bemana2020xfields, stich2008view, zitnick2004high} require multi-view, time synchronized videos as input, and has limited ability to model complicated scene geometry.
Most closely related ours, Yoon~\etal~\cite{yoon2020novel} propose to combine single-view depth and depth from multi-view stereo to render novel views by performing explicit depth based 3D warping.
However, this method has several drawbacks: it relies on human annotated foreground masks, requires cumbersome preprocessing and pretraining and tends to produce artifacts in disocclusions. Instead, we show that our model can be trained end-to-end and produces much more realistic results, and is able to represent complicated scene structure and view dependent effects along with natural degrees of motion.

\paragraph{Dynamic scene reconstruction.}
Most successful non-rigid reconstruction systems either require RGBD data as input~\cite{bozic2020deepdeform, Dou2016Fusion4DRP, Innmann2016VolumeDeformRV, newcombe2015dynamicfusion, ye2014real, zollhofer2014real}, or can only reconstruct sparse geometry~\cite{Park20103DRO, Simon2017KroneckerMarkovPF, Vo2016SpatiotemporalBA, Zheng2015SparseD3}. A few prior monocular methods proposed using strong hand-crafted priors to decompose dynamic scenes into piece-wise rigid parts~\cite{Kumar2017MonocularD3, ranftl2016dense, russell2014video}. Recent work of Luo~\etal~\cite{xuan2020consistent} estimates temporally consistent depth maps of scenes with small object motion by optimizing the weights of a single image depth prediction network, but we show that this approach fails to model large and complex 3D motions. 
Additional work has aimed to predict per-pixel scene flows of dynamic scenes from either monocular or RGBD sequences~\cite{brickwedde2019mono, hur2020self, jiang2019sense, lv2018learning, mittal2020just, sun2015layered}.%

\section{Approach}
We build upon prior work for static scenes~\cite{mildenhall2020nerf}, to which we add the notion of time, and estimate 3D motion by explicitly modeling forward and backward scene flow as dense 3D vector fields.
In this section, we first describe this time-variant (dynamic) scene representation (Sec.~\ref{sec:local}) and the method for effectively optimizing this representation (Sec.~\ref{sec:optimization}) on the input views. We then discuss how to improve the rendering quality by adding an additional explicit time-invariant (static) scene representation, optimized jointly with the dynamic one by combining both during rendering (Sec.~\ref{sec:global}). Finally, we describe how to achieve space-time interpolation of dynamic scenes through our trained representation (Sec.~\ref{sec:space-time}).

\paragraph{Background: static scene rendering.}
Neural Radiance Fields (NeRFs)~\cite{mildenhall2020nerf} represent a static scene as a radiance field defined over a bounded 3D volume. 
This radiance field, denoted $F_\Theta$, is defined by a set of parameters $\Theta$ that are optimized to reconstruct the input views. 
In NeRF, $F_\Theta$ is a multi-layer perceptron (MLP) that takes as
input a position ($\Pos$) and viewing direction ($\vd$), and produces as output a volumetric density
($\sigma$) and RGB color ($\mathbf{c}$):
\begin{align}
        (\colorStatic, \sigma ) = F_{\Theta} (\Pos, \vd )
        \label{eq:static_F}
\end{align}
To render the color of an image pixel, NeRF approximates a volume rendering integral.
Let $\rayStatic$ be the camera ray emitted from the center of projection through a pixel on the image plane. The expected color $\mathbf{\hat{C}}$ of that pixel is then given by:
\begin{align}
    \colorRenderStatic (\rayStatic) = \int_{t_n}^{t_f} T(t) \thinspace \sigma(\rayStatic(t)) \thinspace \colorStatic (\rayStatic(t),\vd) \thinspace dt\nonumber \\
    \text{where \ } T(t) = \exp \left(- \int_{t_n}^t \sigma( \rayStatic (s) ) \thinspace ds \right).
    \label{eq:volren}
\end{align}
Intuitively, $T(t)$ corresponds to the accumulated transparency along that ray. 
The loss is then the difference between the reconstructed color $\mathbf{\hat{C}}$, and the ground truth color $\mathbf{C}$ corresponding to the pixel that ray originated from $\rayStatic$:
\begin{align}
    \Lstatic = \sum_{\rayStatic} || \colorRenderStatic (\rayStatic) - \colorGTStatic (\rayStatic) ||^2_2. \label{eq:static_loss}
\end{align}

\subsection{Neural scene flow fields for dynamic scenes}
\label{sec:local}
To capture scene dynamics, we extend the static scenario described in Eq.~\ref{eq:static_F} by including time in the domain and explicitly modeling 3D motion as dense scene flow fields.
For a given 3D point $\Pos$ and time $i$, the model predicts not just reflectance and opacity, but also forward and backward 3D scene flow $\sfSet = (\sfFwd, \sfBwd)$, which denote 3D offset vectors that point to the position of $\Pos$ at times $i+1$ and $i-1$ respectively.
Note that we make the simplifying assumption that movement that occurs between observed time instances is linear.
To handle motion disocclusions in 3D space, we also predict disocclusion weights $\weightSet = (\wFwd, \wBwd)$ (described in Sec.~\ref{sec:optimization}). 
Our dynamic model is thus defined as:
\begin{align}
       (\colori, \sigma_i, \sfSet, \weightSet ) = F^{\text{dy}}_{\Theta} (\Pos, \vd, i ). \label{eq:dynamic_F}
\end{align}
Note that for convenience, we use the subscript $i$ to indicate a value at a specific time $i$.

\subsection{Optimization}
\label{sec:optimization}

\begin{figure}[t]
    \centering
    \includegraphics[width=\columnwidth]{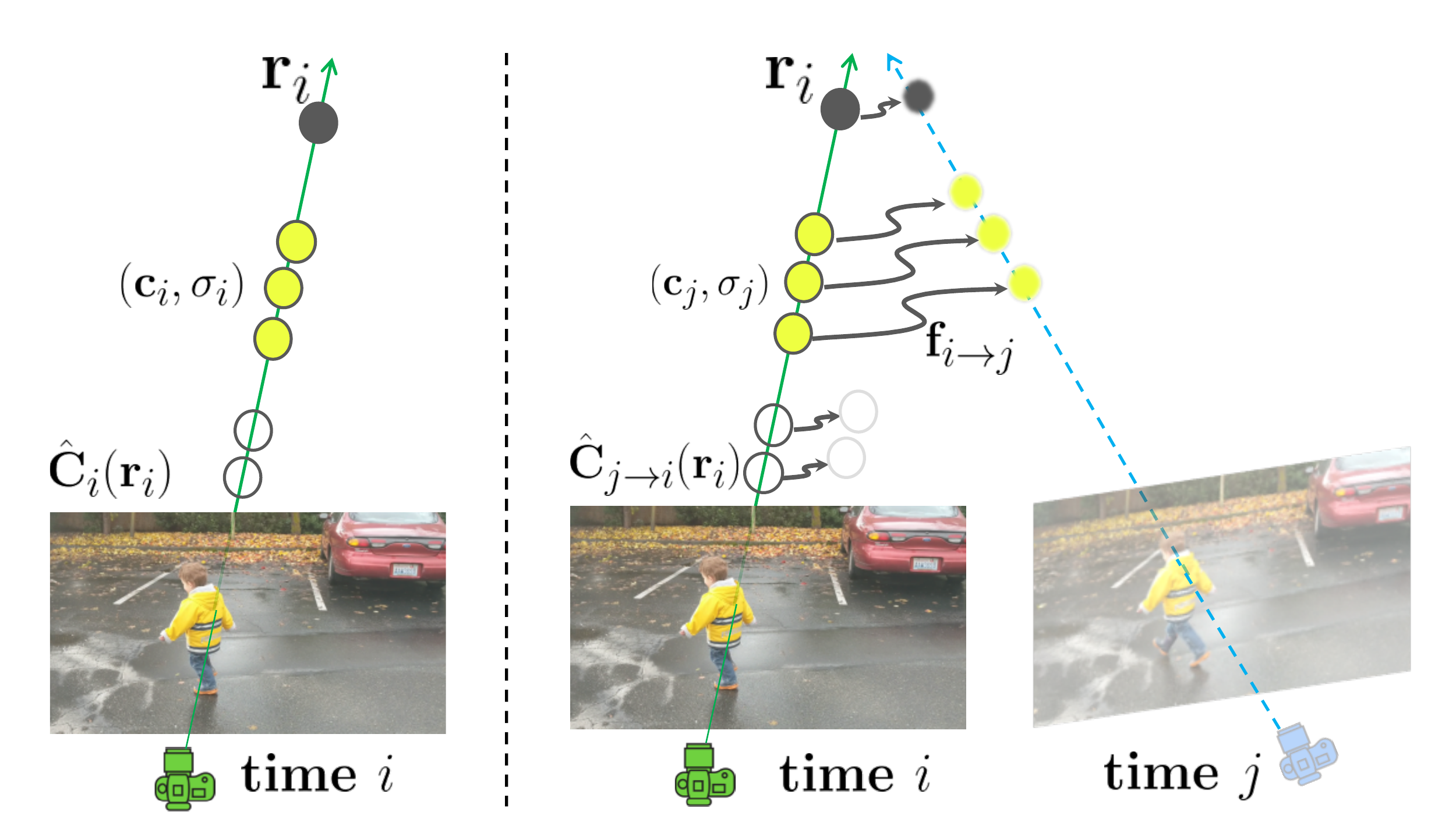}
  \caption{\textbf{Scene flow fields warping.} To render a frame at time $i$, we perform volume rendering along ray $\rayi$ with RGB$\sigma$ at time $i$, giving us the pixel color $\colorRenderStatic_i(\rayi)$ (left). To warp the scene from time $j$ to $i$, we offset each step along $\rayi$ using scene flow $\sfij$ and volume render with the associated color and opacity  $(\colorj, \sigma_j)$ (right).
  \label{fig:temporal_photo_const}}
\end{figure}

\paragraph{Temporal photometric consistency.}
The key new loss we introduce enforces that the scene at time $i$ should be consistent with the scene at neighboring times $j \in \mathcal{N}(i)$, \emph{when accounting for motion that occurs due to 3D scene flow}.
To do this, we volume render the scene at time $i$ from 1) the perspective of the camera at time $i$
and 2) with the scene warped from $j$ to $i$, so as to undo any motion that occurred from $i$ to $j$. 
As shown in Fig.~\ref{fig:temporal_photo_const} (right), during volume rendering, we achieve this by warping each 3D sampled point location $\Posi$ along a ray $\rayi$  using the predicted scene flows fields $\sfSet$ to look up the RGB color $\colorj$ and opacity $\sigma_j$ at neighboring time $j$.
This yields a rendered image, denoted $\colorRenderj$, of the scene at time $j$ with both camera and scene motion warped to time $i$:
\begin{align}
    \colorRenderj (\rayi) & = \int_{t_n}^{t_f} \Tjp (t) \thinspace \sigma_{j} (\rayDisp (t)) \thinspace \mathbf{c}_{j} ( \rayDisp (t), \vdi ) dt  \nonumber \\
    \text{where \ } &
    \rayDisp (t) = \rayi (t) + \sfij ( \rayi (t) ). \label{eq:dynamic_render}
\end{align}
We minimize the mean squared error (MSE) between each warped rendered view and the ground truth view:
\begin{align}
    \Lpho = \sum_{\rayi} \sum_{j \in \mathcal{N}(i) }|| \colorRenderj (\rayi) - \mathbf{C}_i(\rayi) ||_2^2 \label{eq:photo_loss}
\end{align}

\begin{figure}[t]
    \centering
    \includegraphics[width=0.8\columnwidth]{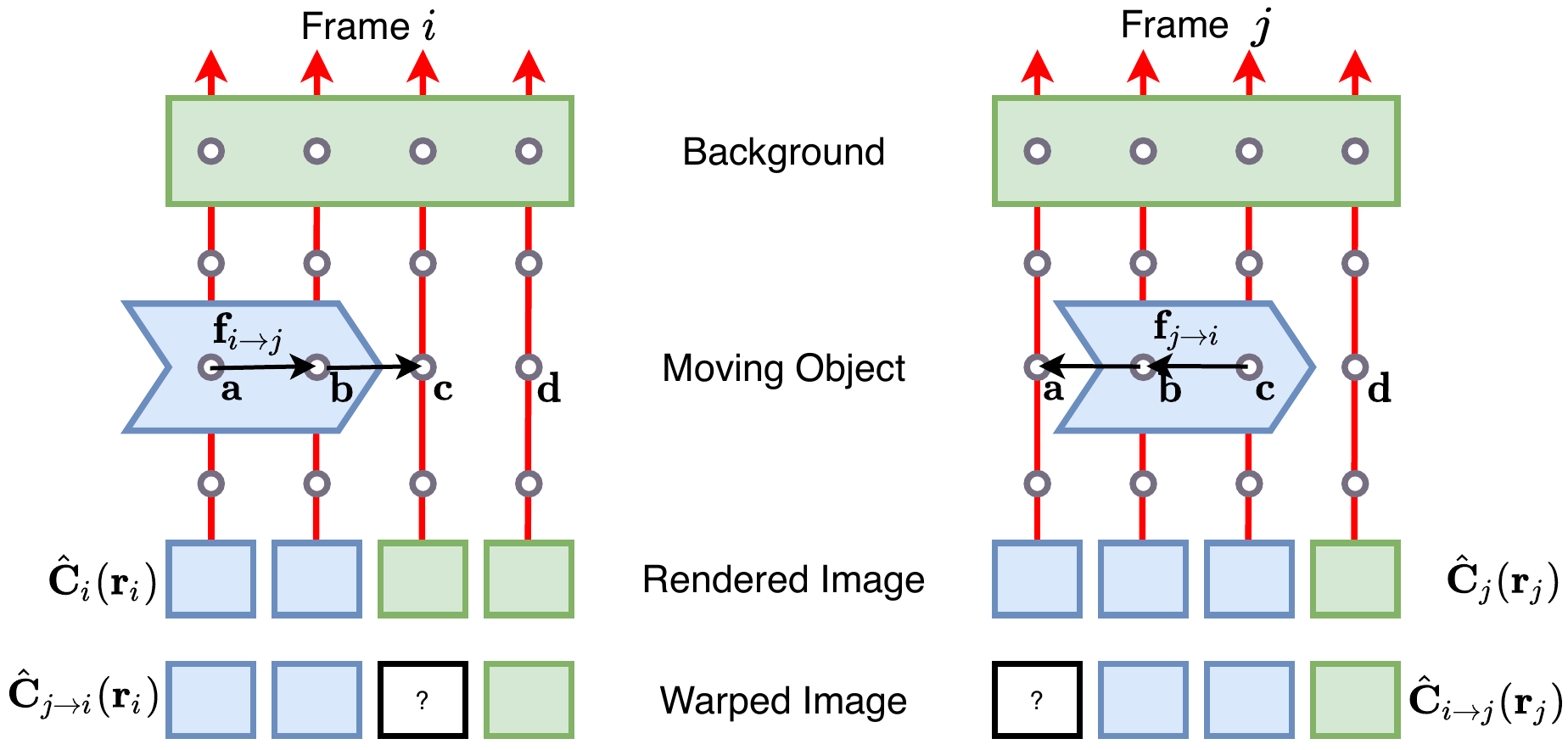}
  \caption{\textbf{Scene flow disocclusion ambiguity.} In this 2D orthographic example, a single blue object translates to the right by one unit from frame $i$ to frame $j$. Here, the correct scene flow at the point labeled $\mathbf{a}$, e.g., $\sfij \left(\mathbf{a} \right)$, points one unit to the right, however, for the scene flow $\sfij \left(\mathbf{c} \right)$ (and similarly $\sfji \left(\mathbf{a}\right)$), there can be multiple answers. If $\sfij \left(\mathbf{c} \right)=0$, then the scene flow would incorrectly point to the foreground in the next frame, and if $\sfij \left(\mathbf{c} \right)=1$, the scene flow would point to the free-space location $d$ at $j$.}
  \label{fig:disocculusion_explain}
\end{figure}

\begin{figure}[t]
    \centering
    \setlength{\tabcolsep}{0.025cm}
    \setlength{\itemheight}{1.8cm}
    \setlength{\itemwidth}{1.23cm}
    \renewcommand{\arraystretch}{0.5}
    \hspace*{-\tabcolsep}\begin{tabular}{ccccc}
            \multirow{2}{*}[0.5\itemheight-5pt]{\begin{tikzpicture}
                \definecolor{boxcolor}{RGB}{255,0,0}
                \node [anchor=south west, inner sep=0.0cm] (image) at (0,0) {
                    \includegraphics[height=\itemheight]{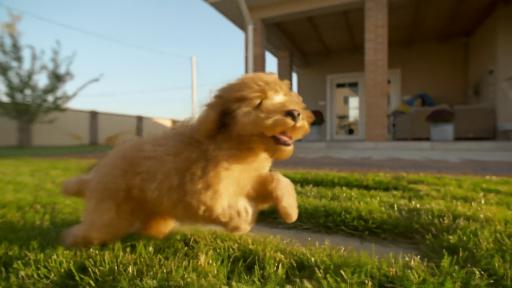}
                };
                \begin{scope}[x={(image.south east)},y={(image.north west)}]
                    \draw [boxcolor] (0.23,0.78) rectangle (0.58,0.37);
                \end{scope}
            \end{tikzpicture}}
        &
            \includegraphics[width=\itemwidth, trim={4.0cm 3.85cm 8.5cm 2.4cm}, clip]{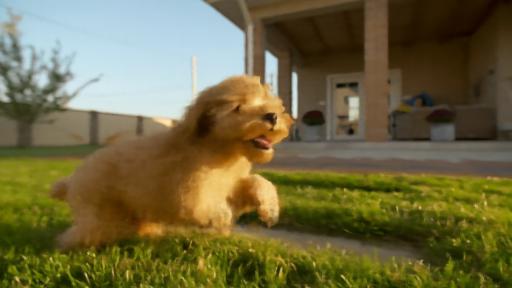}
        &
            \includegraphics[width=\itemwidth, trim={5.0cm 3.85cm 7.5cm 2.4cm}, clip]{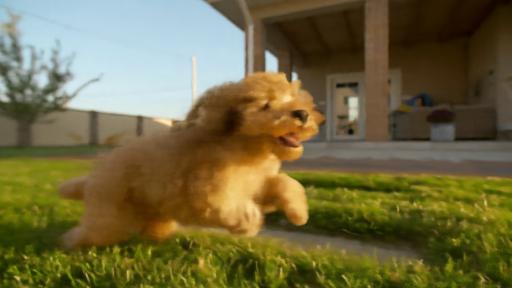}
        &
            \includegraphics[width=\itemwidth, trim={5.0cm 3.85cm 7.5cm 2.4cm}, clip]{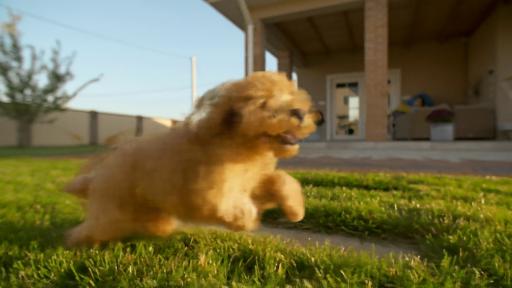}
        &
            \includegraphics[width=\itemwidth, trim={4.8cm 3.85cm 7.7cm 2.4cm}, clip]{figures/weights/w_w/012.jpg}
        \\
            
        &
            \includegraphics[width=\itemwidth, trim={4.0cm 3.85cm 8.5cm 2.4cm}, clip]{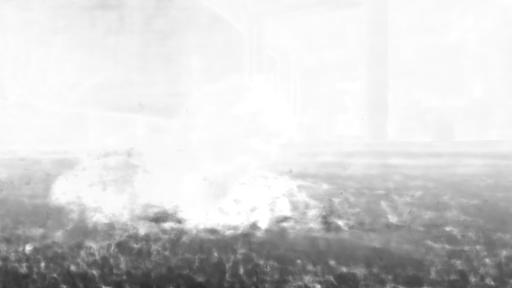}
        &
            \includegraphics[width=\itemwidth, trim={5.0cm 3.85cm 7.5cm 2.4cm}, clip]{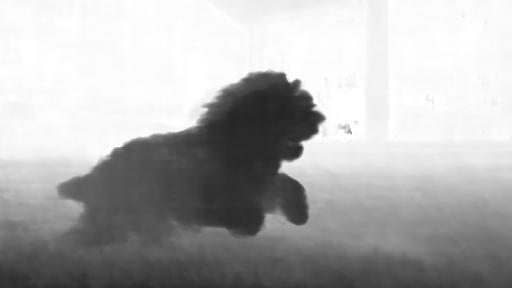}
        &
            \includegraphics[width=\itemwidth, trim={5.0cm 3.85cm 7.5cm 2.4cm}, clip]{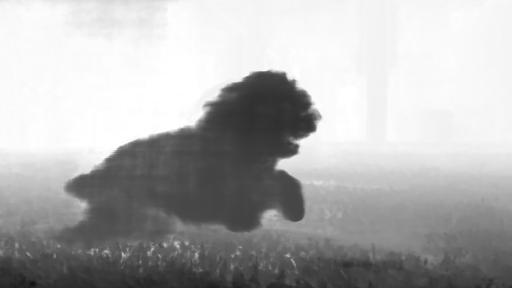}
        &
            \includegraphics[width=\itemwidth, trim={4.8cm 3.85cm 7.7cm 2.4cm}, clip]{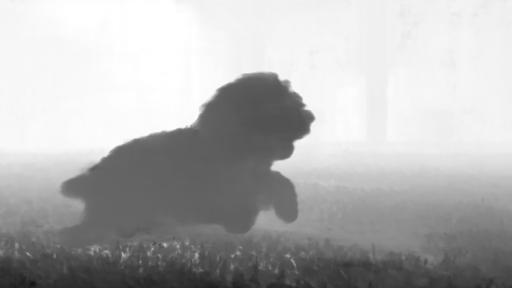}
        \\
            \multirow{2}{*}[0.5\itemheight-5pt]{\begin{tikzpicture}
                \definecolor{boxcolor}{RGB}{255,0,0}
                \node [anchor=south west, inner sep=0.0cm] (image) at (0,0) {
                    \includegraphics[height=\itemheight]{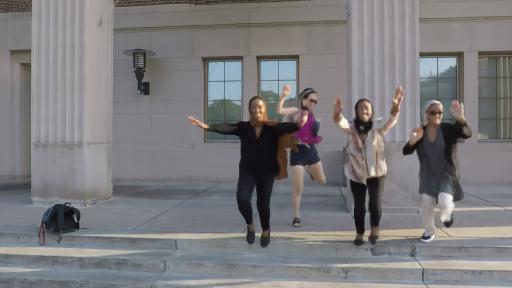}
                };
                \begin{scope}[x={(image.south east)},y={(image.north west)}]
                    \draw [boxcolor] (0.35,0.82) rectangle (0.84,0.2);
                \end{scope}
            \end{tikzpicture}}
        &
            \includegraphics[width=\itemwidth, trim={6.5cm 2.2cm 2.8cm 1.8cm}, clip]{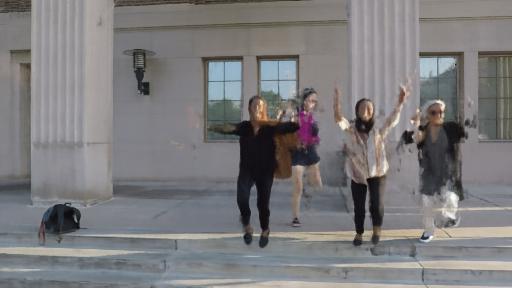}
        &
            \includegraphics[width=\itemwidth, trim={6.5cm 2.2cm 2.8cm 1.8cm}, clip]{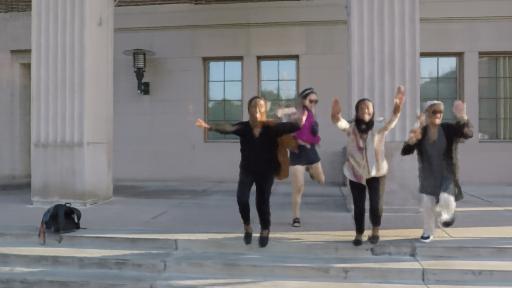}
        &
            \includegraphics[width=\itemwidth, trim={6.5cm 2.2cm 2.8cm 1.8cm}, clip]{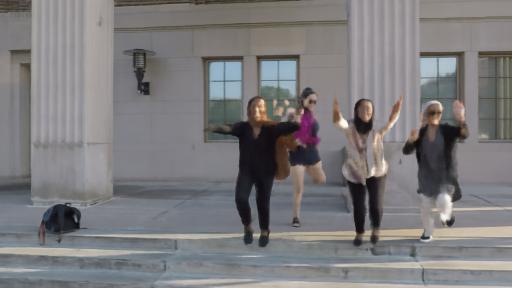}
        &
            \includegraphics[width=\itemwidth, trim={6.5cm 2.2cm 2.8cm 1.8cm}, clip]{figures/weights/w_w/020.jpg}
        \\
            
        &
            \includegraphics[width=\itemwidth, trim={6.5cm 2.2cm 2.8cm 1.8cm}, clip]{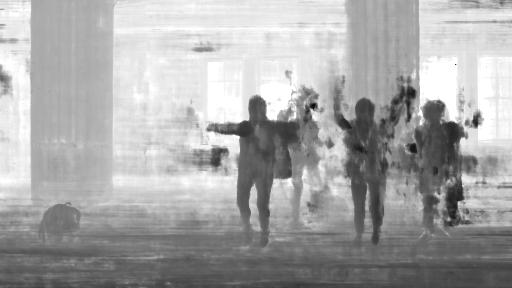}
        &
            \includegraphics[width=\itemwidth, trim={6.5cm 2.2cm 2.8cm 1.8cm}, clip]{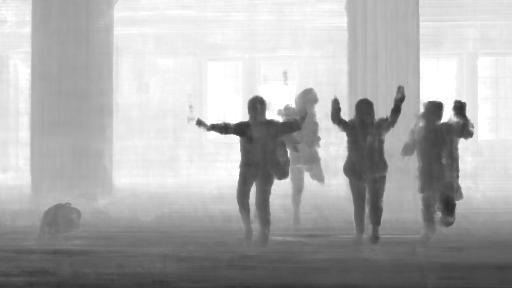}
        &
            \includegraphics[width=\itemwidth, trim={6.5cm 2.2cm 2.8cm 1.8cm}, clip]{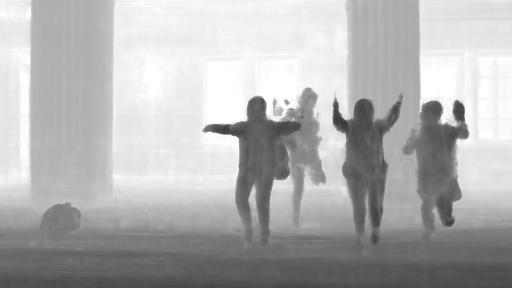}
        &
            \includegraphics[width=\itemwidth, trim={6.5cm 2.2cm 2.8cm 1.8cm}, clip]{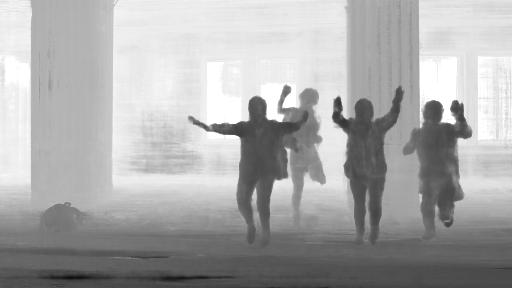}
        \\
            \vphantom{\footnotesize I}\tiny (a) Our rendered views
        &
            \vphantom{\footnotesize I}\tiny (b) w/o $\Ldata$
        &
            \vphantom{\footnotesize I}\tiny (c) w/o $\Lcycle$
        &
            \vphantom{\footnotesize I}\tiny (d) w/o $\weightSet$
        &
            \vphantom{\footnotesize I}\tiny (e) Full
        \\
    \end{tabular}\vspace{-0.2cm}
    \caption{\textbf{Qualitative ablations.} Results of our full method with different loss components removed. The odd rows show zoom-in rendered color and the even rows show corresponding pseudo depth. Each component reduces the overall quality in different ways.}
  \label{fig:ablation_loss}
\end{figure}

An important caveat is that this loss is ambiguous at 3D disocculusion regions caused by motion. 
Analogous to 2D dense optical flow~\cite{meister2017unflow}, scene flow is ambiguous when a 3D location becomes occluded or disoccluded between frames.
These regions are especially important as they occur at the boundaries of moving objects (see Fig.~\ref{fig:disocculusion_explain} for an illustration).
To mitigate errors due to this ambiguity, we predict extra continuous disocclusion weight fields $\wFwd$ and $\wBwd \in [0, 1 ]$, corresponding to $\sfFwd$ and $\sfBwd$ respectively. 
These weights serve as an unsupervised confidence of where and how much strength the temporal photoconsistency loss should be applied.
We apply these weights by volume rendering the weight along the ray $\rayi$ with opacity from time $j$, and multiplying the accumulated weight at each \emph{2D pixel}:%
\begin{align}
    \wRender (\rayi) =  \int_{t_n}^{t_f} \funop{T_j(t)} \funop{\sigma_{j} (\rayDisp (t))} \funop{\wij (\rayi (t))} \, dt \label{eq:render_w}
\end{align}
We avoid the trivial solution where all predicted weights are zero by adding $\ell_1$ regularization to encourage predicted weights to be close to one, giving us a new weighted loss:
\begin{align}
    \Lpho = \sum_{\rayi} \sum_{j \in \mathcal{N}(i)} & \wRender (\rayi) || \colorRenderj (\rayi) - \mathbf{C}_i(\rayi) ||_2^2 \nonumber \\
    & + \Wdisocc \sum_{\Posi}|| \wij (\Posi) - 1\||_1, \label{eq:weightedtemporalphotometricloss}
\end{align}
where $\Wdisocc$ is a regularization weight which we set to $0.1$ in all our experiments. 
We use $\mathcal{N}(i) = \{i, i \pm 1, i \pm 2 \}$, and chain scene flow and disocclusion weights for the $i \pm 2$ cases.
Note that when $j=i$, there is no scene flow warping or disocculusion weights involved ($\sfij=0, \wRender (\rayi) = 1)$, meaning that $\colorRenderii (\rayi) = \colorRenderi (\rayi)$, as in Fig.~\ref{fig:temporal_photo_const}(left).
Comparing Fig.~\ref{fig:ablation_loss}(e) and Fig.~\ref{fig:ablation_loss}(d), we can see that adding this disocclusion weight improves rendering quality near motion boundaries. %

\paragraph{Scene flow priors.}
To regularize the predicted scene flow fields, we add a 3D scene flow cycle consistency term to encourage that at all sampled 3D points $\Posi$, the predicted forward scene flow $\sfij$ is consistent with the backward scene flow $\sfji$ at the corresponding location at time $j$ (i.e. at position $\Posij = \Posi + \sfij$).
Note that this cycle consistency can also be ambiguous near motion disocclusion regions in 3D, so we use the same predicted disocclusion weights to modulate this term, giving us:
{
\begin{align}
    \Lcycle = \sum_{\Posi} \sum_{j \in i \pm 1} \wij || \sfij (\Posi)  + \sfji (\Posij)||_1 \label{eq:cycle_loss}
\end{align}
}

We additionally add low-level regularizations $\Lphy$ on the predicted scene flow fields.  
First, following prior work~\cite{newcombe2015dynamicfusion, Vo2016SpatiotemporalBA}, we enforce scene flow spatial-temporal smoothness by applying $\ell_1$ regularization to nearby sampled 3D points along the ray and encouraging 3D point trajectories to be piece-wise linear. Second, we encourage scene flow to be small in most places~\cite{Valmadre2012GeneralTP} by applying an $\ell_1$ regularization term. 
Please see the supplementary material for complete descriptions.

\paragraph{Data-driven priors.}
Since monocular reconstruction of complex dynamic scenes is highly ill-posed, the above losses can on occasion converge to sub-optimal local minima when randomly initialized. 
Therefore, we introduce two data-driven losses, a geometric consistency term~\cite{Innmann2016VolumeDeformRV, Innmann2020NRMVSNM} and a single-view depth term: $\Ldata = \Lgeo + \Wdisp \Ldisp$. We set $\Wdisp = 2$ in all our experiments.

The geometric consistency helps model build accurate correspondence association between adjacent frames. In particular, it minimizes the reprojection error of scene flow displaced 3D points w.r.t. the derived 2D optical flow which we compute using pretrained networks~\cite{ilg2017flownet, teed2020raft}. 

Suppose $\pixel$ is a 2D pixel position at time $i$. The corresponding 2D pixel location in the neighboring frame at time $j$ displaced through 2D optical flow $\ofij$ can be computed as $\pixelDisp = \pixel + \ofij$. To estimate the expected 2D point location $\pixelExp$ at time $j$ displaced by predicted scene flow fields, we first compute the expected scene flow $\sfExpij (\rayi)$ and the expected 3D point location $\epl (\rayi)$ of the ray $\rayi$ through volume rendering.
$\pixelExp$ is then computed by performing perspective projection of the expected 3D point location displaced by the scene flow (i.e. $\epl (\rayi) + \sfExpij(\rayi)$) into the viewpoint corresponding to the frame at time $j$. 
The geometric consistency is computed as the $\ell_1$ difference between $\pixelExp$ and $\pixelDisp$,
\begin{align}
    \Lgeo = \sum_{\rayi} \sum_{j \in \{i \pm 1\}}  || \pixelExp (\rayi)- \pixelDisp (\rayi))||_1.
\end{align}

We also add a single view depth prior that encourages the expected termination depth $\depthRender_i$ computed along each ray to be close to the depth $\depthSV_i$ predicted from a pre-trained single-view network~\cite{Ranftl2020}. 
As single-view depth predictions are defined up to an unknown scale and shift, we utilize a robust scale-shift invariant loss~\cite{Ranftl2020}:
\begin{align}
\Ldisp = \sum_{\rayi} || \depthRender_i^* (\rayi)- \depthSV_i^* (\rayi)||_1
\end{align}
where ${}^{*}$ is a whitening operation that normalizes the depth to have zero mean and unit scale.

From Fig~\ref{fig:ablation_loss}(b), we see that adding data-driven priors help the model learn correct scene geometry especially for dynamic regions.
However, as both of these data-driven priors are noisy (rely on inaccurate or incorrect predictions), we use these for \emph{initialization only}, and linearly decay the weight of $\Ldata$ to zero during training.

\begin{figure}[t]
    \centering
    \setlength{\tabcolsep}{0.025cm}
    \setlength{\itemwidth}{2.73cm}
    \renewcommand{\arraystretch}{0.5}
    \hspace*{-\tabcolsep}\begin{tabular}{ccc}
            \includegraphics[width=\itemwidth]{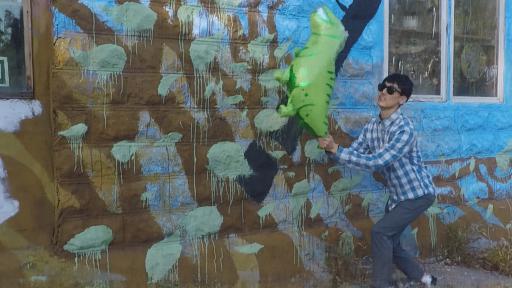}
        &
            \includegraphics[width=\itemwidth]{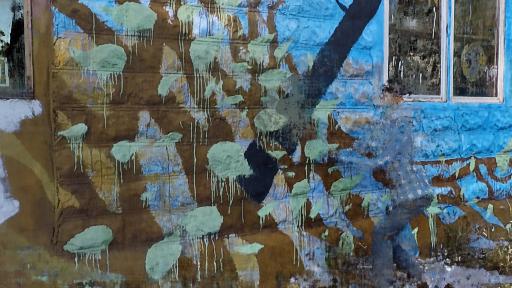}
        &
            \includegraphics[width=\itemwidth]{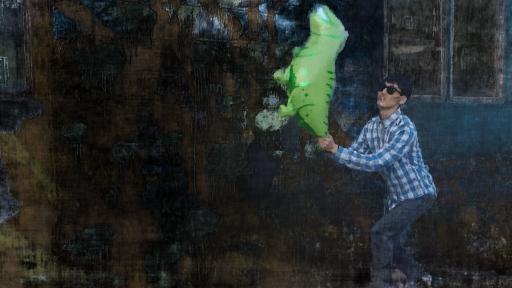}
        \\
            \includegraphics[width=\itemwidth]{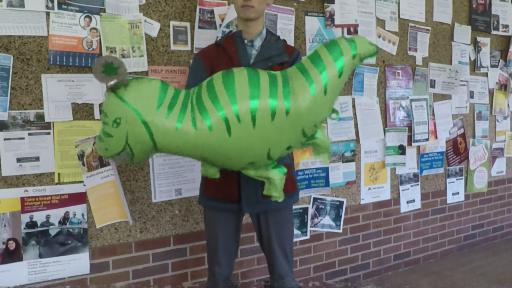}
        &
            \includegraphics[width=\itemwidth]{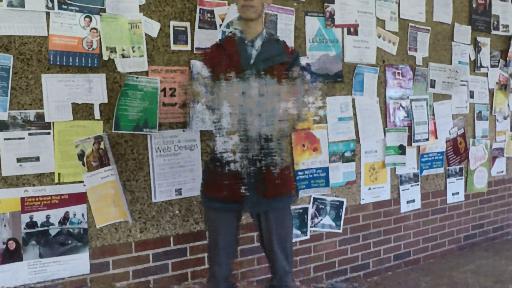}
        &
            \includegraphics[width=\itemwidth]{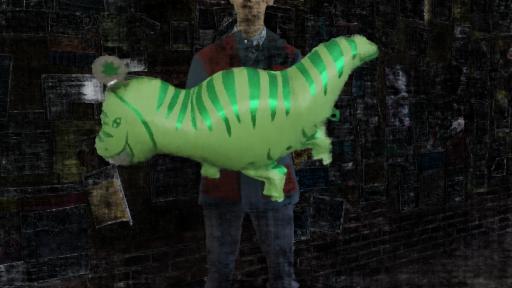}
        \\
            \vphantom{I}\footnotesize Combined render
        &
            \vphantom{I}\footnotesize Static only
        &
            \vphantom{I}\footnotesize Dynamic only
        \\
    \end{tabular}\vspace{-0.2cm}
    \caption{\textbf{Dynamic and static components.} Our method learns static and dynamic components in the combined representation. Note person is almost still in the bottom example.}
    \label{fig:render_decomposition}
\end{figure}

\subsection{Integrating a static scene representation}
\label{sec:global}
\begin{figure}[t]
    \centering
    \setlength{\tabcolsep}{0.025cm}
    \setlength{\itemheight}{2.2cm}
    \setlength{\itemwidth}{2.15cm}
    \renewcommand{\arraystretch}{0.5}
    \hspace*{-\tabcolsep}\begin{tabular}{ccc}
            \multirow{2}{*}[0.5\itemheight-5pt]{\begin{tikzpicture}
                \definecolor{boxcolor}{RGB}{255,0,0}
                \node [anchor=south west, inner sep=0.0cm] (image) at (0,0) {
                    \includegraphics[height=\itemheight]{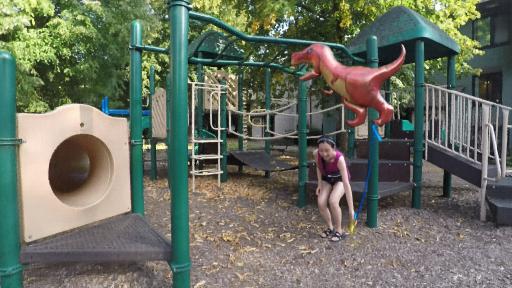}
                };
                \begin{scope}[x={(image.south east)},y={(image.north west)}]
                    \draw [boxcolor] (0.35,0.96) rectangle (0.92,0.43);
                \end{scope}
            \end{tikzpicture}}
        &
            \includegraphics[width=\itemwidth, trim={6.4cm 4.5cm 1.0cm 0.36cm}, clip]{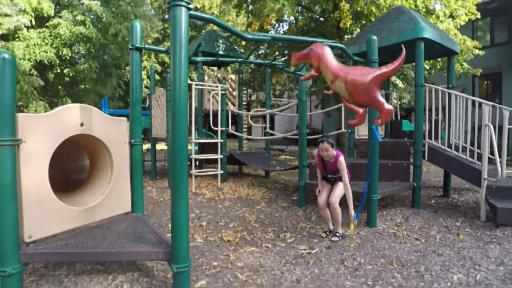}
        &
            \includegraphics[width=\itemwidth, trim={6.4cm 4.5cm 1.0cm 0.36cm}, clip]{figures/global/w/000.jpg}
        \\
            
        &
            \includegraphics[width=\itemwidth, trim={6.4cm 4.5cm 1.0cm 0.36cm}, clip]{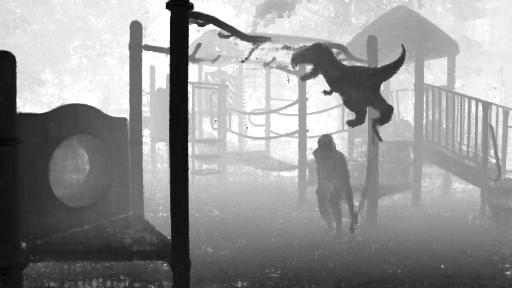}
        &
            \includegraphics[width=\itemwidth, trim={6.4cm 4.5cm 1.0cm 0.36cm}, clip]{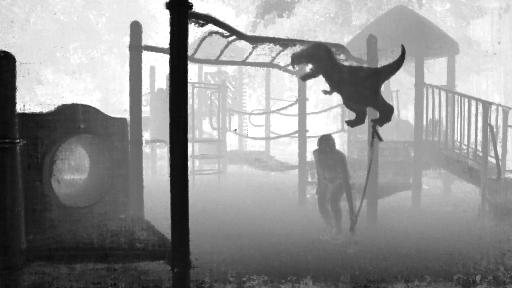}
        \\
            \multirow{2}{*}[0.5\itemheight-5pt]{\begin{tikzpicture}
                \definecolor{boxcolor}{RGB}{255,0,0}
                \node [anchor=south west, inner sep=0.0cm] (image) at (0,0) {
                    \includegraphics[height=\itemheight]{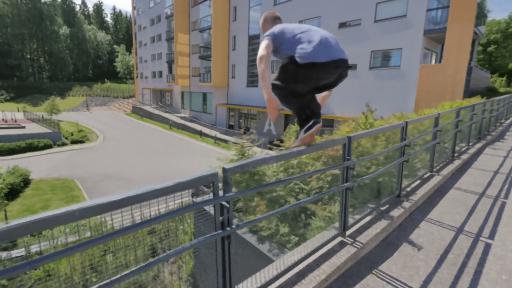}
                };
                \begin{scope}[x={(image.south east)},y={(image.north west)}]
                    \draw [boxcolor] (0.09,0.45) rectangle (0.5,0.1);
                \end{scope}
            \end{tikzpicture}}
        &
            \includegraphics[width=\itemwidth, trim={1.2cm 0.5cm 8.8cm 5.65cm}, clip]{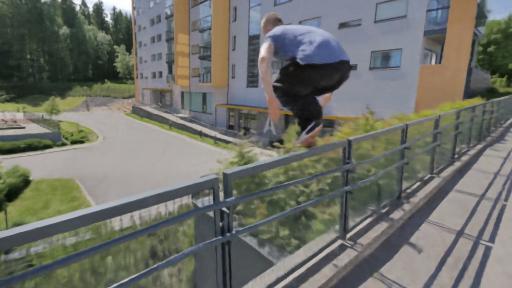}
        &
            \includegraphics[width=\itemwidth, trim={1.2cm 0.5cm 8.8cm 5.65cm}, clip]{figures/global/w/001.jpg}
        \\
            
        &
            \includegraphics[width=\itemwidth, trim={1.2cm 0.5cm 8.8cm 5.65cm}, clip]{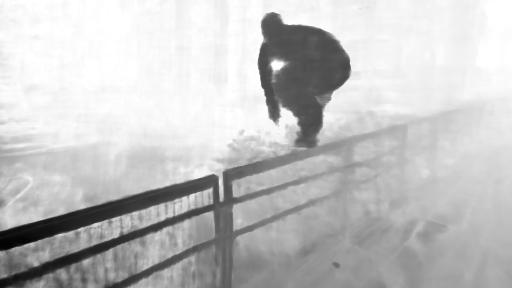}
        &
            \includegraphics[width=\itemwidth, trim={1.2cm 0.5cm 8.8cm 5.65cm}, clip]{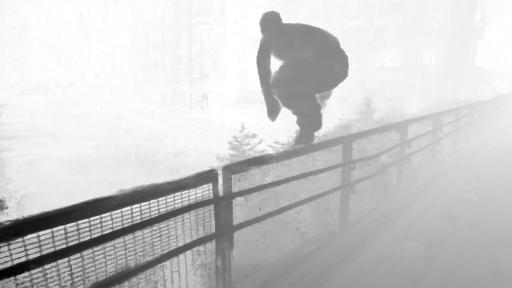}
        \\
            \vphantom{\small I}\scriptsize (a) Our rendered views w/ static
        &
            \vphantom{\small I}\scriptsize (b) w/o static
        &
            \vphantom{\small I}\scriptsize (c) w/ static
        \\
    \end{tabular}\vspace{-0.2cm}
    \caption{\textbf{Static scene representation ablation.} Adding a static scene representation yields higher fidelity renderings, especially in static regions (a,c) when compared to the pure dynamic model (b).}
  \label{fig:global_ablation}
\end{figure}

The method described so far already outperforms the state of the art, as shown in Tab.~\ref{exp:nvidia_quan}.
However, unlike NeRF, our warping-based temporal loss can only be used in a \emph{local} temporal neighborhood $\mathcal{N}(i)$, as dynamic components typically undergo too much deformation to reliably infer correspondence over larger temporal gaps.
Static regions, however, should be consistent and should leverage observations from \emph{all} frames. 
Therefore, we propose to combine our dynamic (time-dependent) scene representation with a static (time-independent) one, and require that when combined, the resulting volume-rendered images match the input. We model each representation with its own MLP, where the dynamic scene component is represented with Eq.~\ref{eq:dynamic_F}, and the static one is represented as a variant of Eq.~\ref{eq:static_F}:
\begin{align}
        (\colorStatic, \sigma, \blendw)= F^{\text{st}}_{\Theta} (\Pos, \vd)  \label{eq:static_F2}
\end{align}
where $\blendw$ is an unsupervised 3D blending weight field, that linearly blends the RGB$\sigma$ from static and dynamic scene representations along the ray. Intuitively, $\blendw$ should assign a low weight to the dynamic representation at static regions with sufficient observations, as these can be rendered in higher fidelity by the static representation, while assigning a lower weight to the static representation in regions that are moving, as these can be better modeled by the dynamic representation. We found adding the extra $\blendw$ leads to less artifacts and more stable training than the configuration without $\blendw$.
The combined rendering equation is then written as:
\begin{align}
    \colorRenderUnionj (\rayi)  = \int_{t_n}^{t_f} \TUnionj (t) \thinspace \opacityUnion_{i} (t) \thinspace \colorjv (t) dt, \label{eq:union_render}
\end{align}
where $\opacityUnion_{i} (t) \thinspace \colorjv (t)$ is a linear combination of static and dynamic scene components, weighted by $\blendw (t)$:
\begin{align}
     \opacityUnion_{i}(t) \thinspace \colorjv(t) = 
      \blendw(t) \thinspace \colorStatic(t) \thinspace \sigma(t) + (1 \hspace{0.03cm} \text{-} \hspace{0.03cm} \blendw(t)) \thinspace \colori(t) \thinspace \sigma_i(t)
\end{align}
For clarity, we omit $\rayi$ in each prediction. 
We then train the combined scene representation by minimizing MSE between $\colorRenderUnionj$ with the corresponding input view:
\begin{align}
    \Lunion = \sum_{\rayi} || \colorRenderUnionj (\rayi) - \colorGTStatic_i (\rayi) ||^2_2.
\end{align}
This loss is added to the previously defined losses on the dynamic representation, giving us the final combined loss: 
\begin{align}
    \mathcal{L} = \Lunion + \Lpho + \Wcycle \Lcycle + \Wdata \Ldata + \Wreg \Lphy
\end{align}
where the $\beta$ coefficients weight each term. 
Fig.~\ref{fig:render_decomposition} shows separately rendered static and dynamic scene components, and Fig.~\ref{fig:global_ablation} visually compares renderings with and without integrating a static scene representation.

\subsection{Space-time view synthesis} \label{sec:space-time}
\begin{figure}[t]
    \centering
    \setlength{\tabcolsep}{0.025cm}
    \setlength{\itemwidth}{2.73cm}
    \renewcommand{\arraystretch}{0.5}
    \hspace*{-\tabcolsep}\begin{tabular}{ccc}
            \begin{tikzpicture}
                \definecolor{boxcolor}{RGB}{255,0,0}
                \node [anchor=south west, inner sep=0.0cm] (image) at (0,0) {
                    \includegraphics[width=\itemwidth]{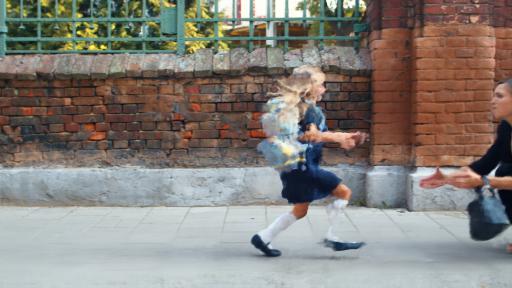}
                };
                \begin{scope}[x={(image.south east)},y={(image.north west)}]
                    \draw [boxcolor] (0.48,0.80) rectangle (0.73,0.08);
                \end{scope}
            \end{tikzpicture}
        &
            \begin{tikzpicture}
                \definecolor{boxcolor}{RGB}{255,0,0}
                \node [anchor=south west, inner sep=0.0cm] (image) at (0,0) {
                    \includegraphics[width=\itemwidth]{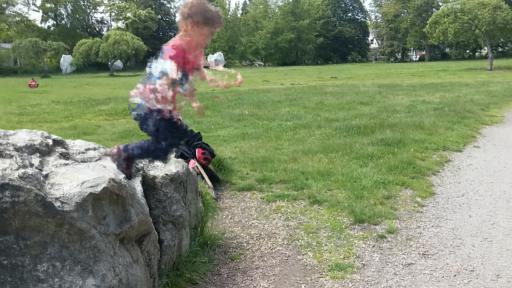}
                };
                \begin{scope}[x={(image.south east)},y={(image.north west)}]
                    \draw [boxcolor] (0.2,0.99) rectangle (0.49,0.36);
                \end{scope}
            \end{tikzpicture}
        &
            \begin{tikzpicture}
                \definecolor{boxcolor}{RGB}{255,0,0}
                \node [anchor=south west, inner sep=0.0cm] (image) at (0,0) {
                    \includegraphics[width=\itemwidth]{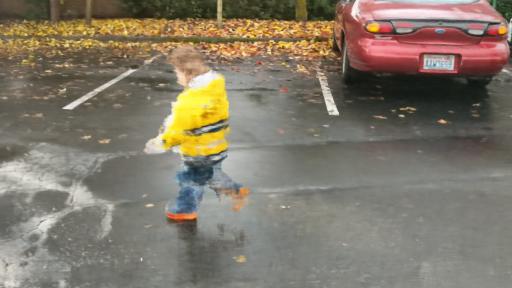}
                };
                \begin{scope}[x={(image.south east)},y={(image.north west)}]
                    \draw [boxcolor] (0.26,0.87) rectangle (0.52,0.2);
                \end{scope}
            \end{tikzpicture}
        \\
            \includegraphics[width=\itemwidth]{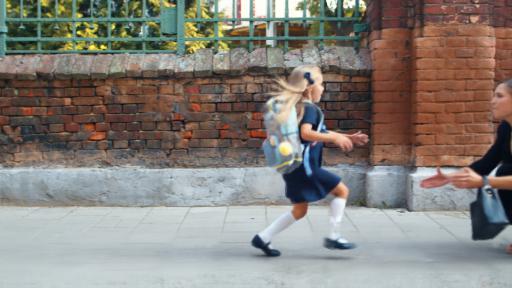}
        &
            \includegraphics[width=\itemwidth]{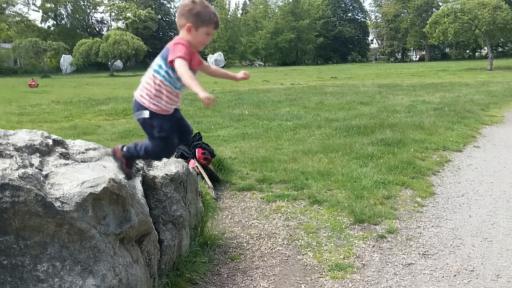}
        &
            \includegraphics[width=\itemwidth]{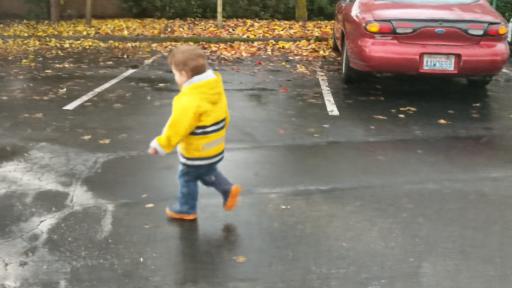}
        \\
    \end{tabular}\vspace{-0.2cm}
    \caption{\textbf{Novel time synthesis.} Rendering images by interpolating the time index (top) yields blending artifacts compared to our scene flow based rendering (bottom).}
  \label{fig:spacetimeinterpolation}
\end{figure}
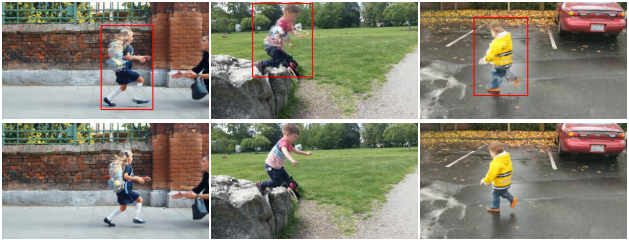

To render novel views at a given time, we simply volume render each pixel using Eq.~\ref{eq:dynamic_render} (dynamic) or Eq.~\ref{eq:union_render} (static+dynamic). 
However, we observe that while this approach produces good results \emph{at times corresponding to input views}, the representation does not allow us to interpolate time-variant geometry at in-between times, leading instead to rendered results that look like linearly blended combinations of existing frames (Fig.~\ref{fig:spacetimeinterpolation}).

Instead, we render intermediate times by warping the scene based on the predicted scene flow. 
For efficient rendering, we propose a splatting-based plane-sweep volume rendering approach. 
To render an image at intermediate time $i + \delta_i, \delta_i\in (0,1)$ at a specified target viewpoint, we sweep over every step emitted from the novel viewpoint from front to back.
At each sampled step $t$ along the ray, we query point information through our models at both times $i$ and $i + 1$, and displace the 3D points at time $i$ by the scaled scene flow $\Posi + \delta_i\sfFwd(\Posi)$, and similarity for time $i+1$.
We then splat the 3D displaced points onto a ($\colorStatic, \alpha$) accumulation buffer at the novel viewpoint, and blend splats from time $i$ and $i+1$ with linear weights $1-\delta_i,\delta_i$.
The final rendered view is obtained by volume rendering the accumulation buffer (see supplementary material for a diagram).

\section{Experiments}

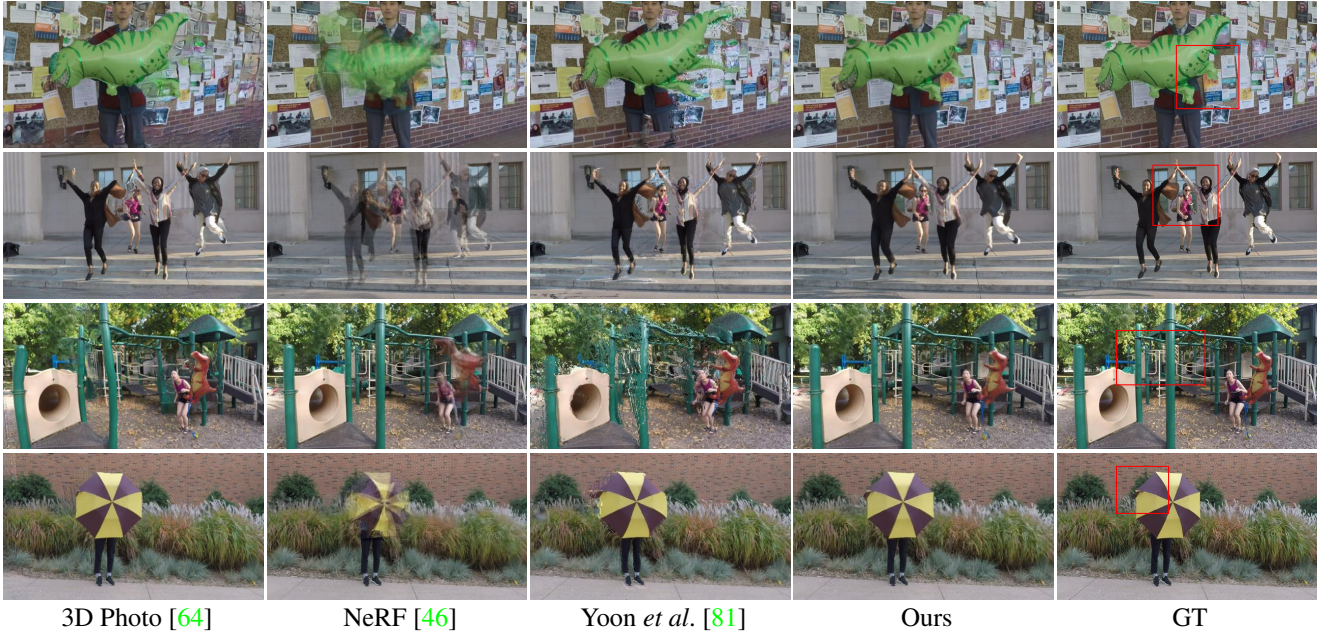
\begin{figure*}[t]
    \centering
    \setlength{\tabcolsep}{0.025cm}
    \setlength{\itemwidth}{3.45cm}
    \renewcommand{\arraystretch}{0.5}
    \hspace*{-\tabcolsep}\begin{tabular}{ccccc}
            \includegraphics[width=\itemwidth]{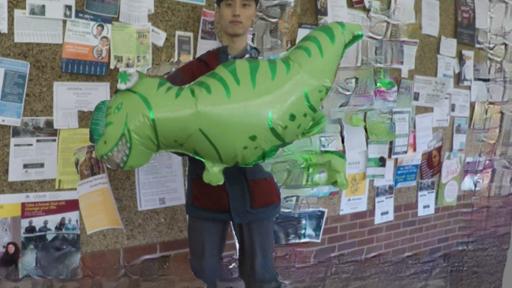}
        &
            \includegraphics[width=\itemwidth]{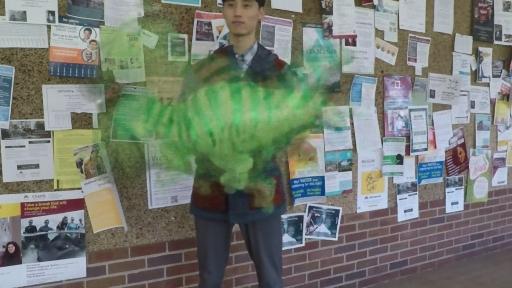}
        &
            \includegraphics[width=\itemwidth]{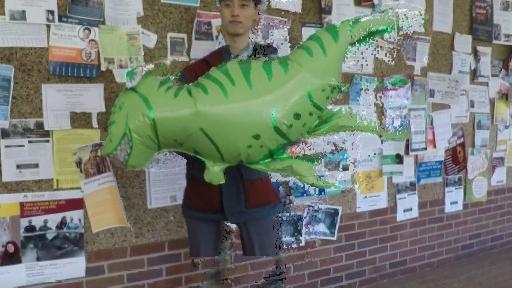}
        &
            \includegraphics[width=\itemwidth]{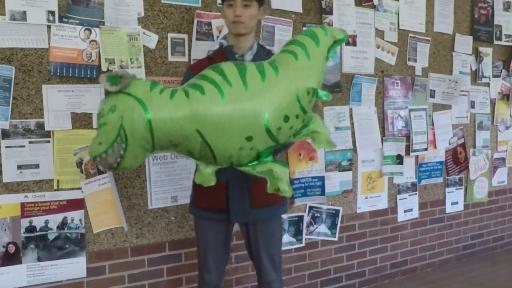}
        &
            \begin{tikzpicture}
                \definecolor{boxcolor}{RGB}{255,0,0}
                \node [anchor=south west, inner sep=0.0cm] (image) at (0,0) {
                    \includegraphics[width=\itemwidth]{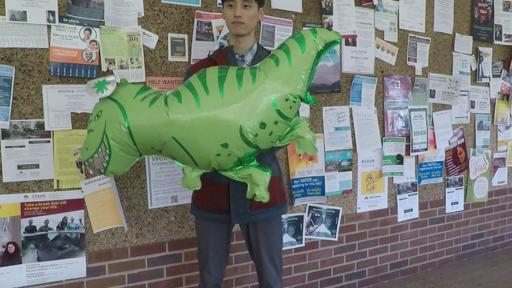}
                };
                \begin{scope}[x={(image.south east)},y={(image.north west)}]
                    \draw [boxcolor] (0.46,0.7) rectangle (0.7,0.27);
                \end{scope}
            \end{tikzpicture}
        \\
            \includegraphics[width=\itemwidth]{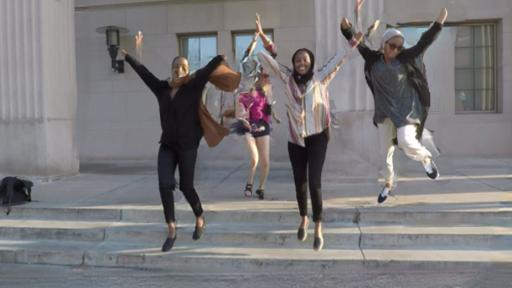}
        &
            \includegraphics[width=\itemwidth]{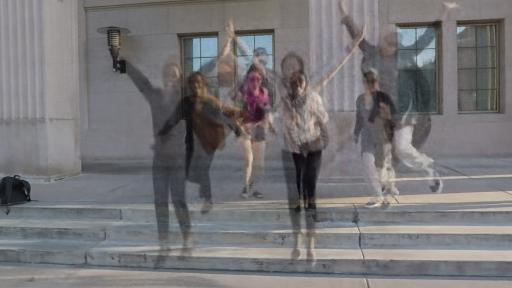}
        &
            \includegraphics[width=\itemwidth]{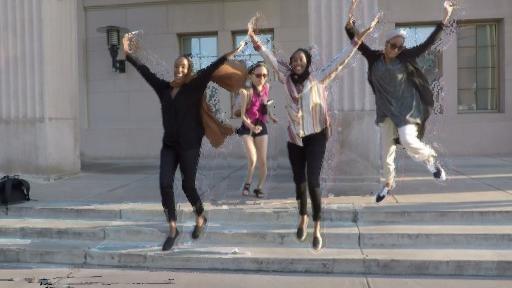}
        &
            \includegraphics[width=\itemwidth]{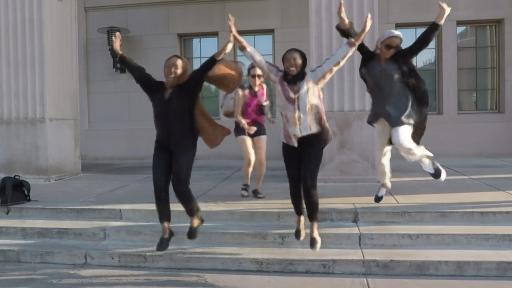}
        &
            \begin{tikzpicture}
                \definecolor{boxcolor}{RGB}{255,0,0}
                \node [anchor=south west, inner sep=0.0cm] (image) at (0,0) {
                    \includegraphics[width=\itemwidth]{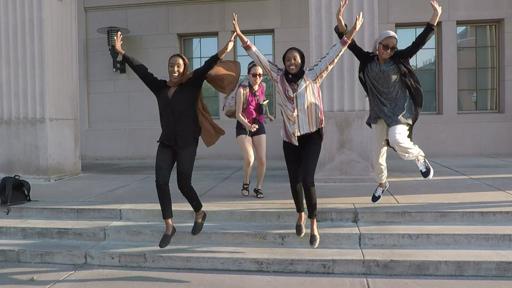}
                };
                \begin{scope}[x={(image.south east)},y={(image.north west)}]
                    \draw [boxcolor] (0.37,0.91) rectangle (0.62,0.5);
                \end{scope}
            \end{tikzpicture}
        \\
            \includegraphics[width=\itemwidth]{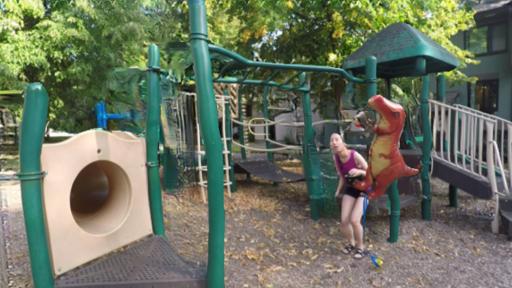}
        &
            \includegraphics[width=\itemwidth]{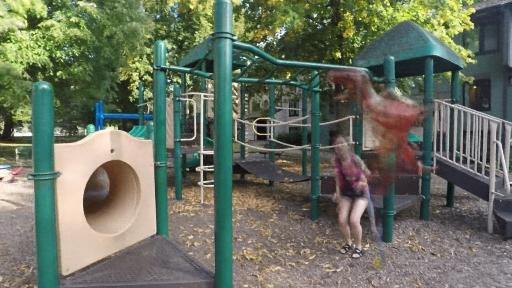}
        &
            \includegraphics[width=\itemwidth]{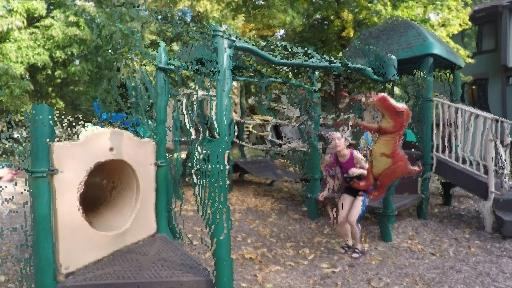}
        &
            \includegraphics[width=\itemwidth]{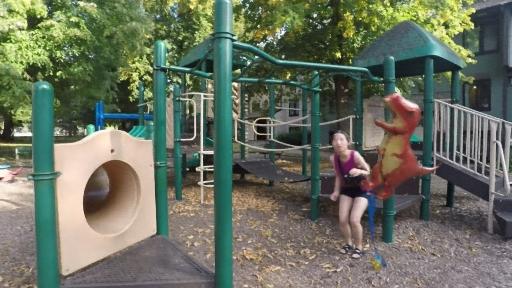}
        &
            \begin{tikzpicture}
                \definecolor{boxcolor}{RGB}{255,0,0}
                \node [anchor=south west, inner sep=0.0cm] (image) at (0,0) {
                    \includegraphics[width=\itemwidth]{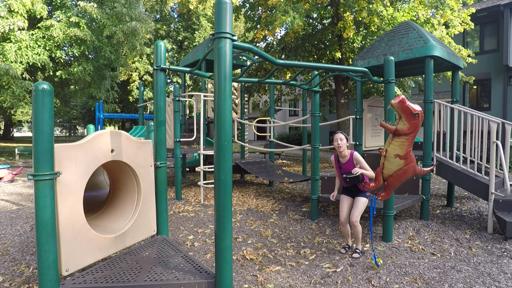}
                };
                \begin{scope}[x={(image.south east)},y={(image.north west)}]
                    \draw [boxcolor] (0.23,0.81) rectangle (0.57,0.44);
                \end{scope}
            \end{tikzpicture}
        \\
            \includegraphics[width=\itemwidth]{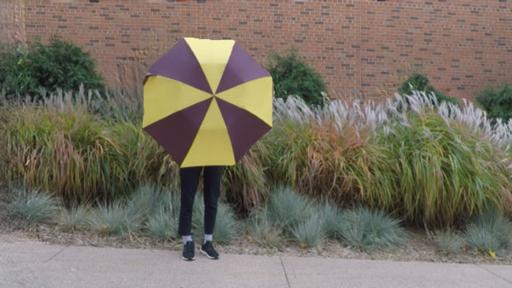}
        &
            \includegraphics[width=\itemwidth]{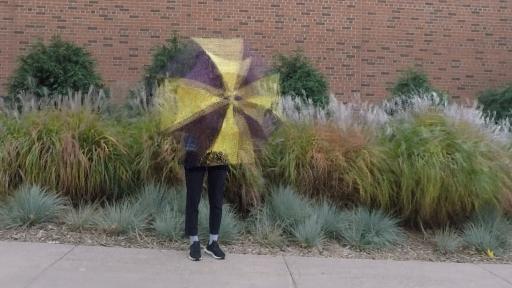}
        &
            \includegraphics[width=\itemwidth]{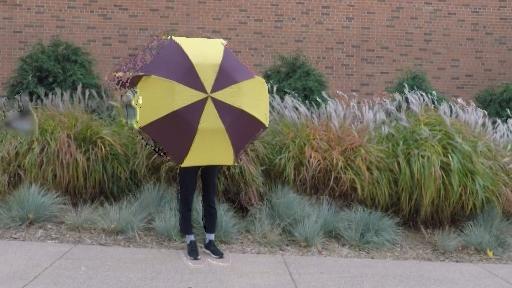}
        &
            \includegraphics[width=\itemwidth]{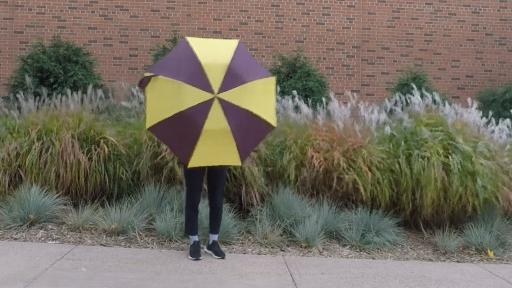}
        &
            \begin{tikzpicture}
                \definecolor{boxcolor}{RGB}{255,0,0}
                \node [anchor=south west, inner sep=0.0cm] (image) at (0,0) {
                    \includegraphics[width=\itemwidth]{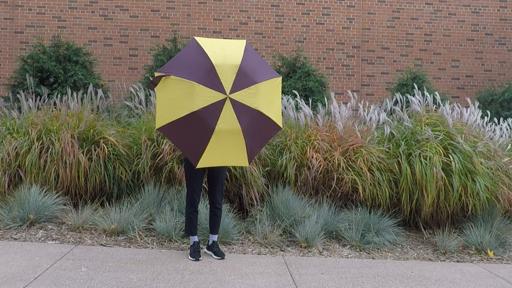}
                };
                \begin{scope}[x={(image.south east)},y={(image.north west)}]
                    \draw [boxcolor] (0.23,0.91) rectangle (0.43,0.59);
                \end{scope}
            \end{tikzpicture}
        \\
            \vphantom{\large I} 3D Photo~\cite{shih20203d}
        &
            \vphantom{\large I} NeRF~\cite{mildenhall2020nerf}
        &
            \vphantom{\large I} Yoon~\etal~\cite{yoon2020novel}
        &
            \vphantom{\large I} Ours
        &
            \vphantom{\large I} GT
        \\
    \end{tabular}\vspace{-0.2cm}
    \caption{\textbf{Qualitative comparisons on the Dynamic Scenes dataset.} Compared with prior methods, our rendered images more closely match the ground truth, and include fewer artifacts, as shown in the highlighted regions.}\vspace{-0.2cm}
    \label{fig:nvidia_qual}
\end{figure*}

\paragraph{Implementation details.} 
We use COLMAP~\cite{schonberger2016structure} to estimate camera intrinsics and extrinsics, and consider these fixed during optimization.
As COLMAP assumes a static scene, we mask out features from regions associated with common classes of dynamic objects using off-the-shelf instance segmentation~\cite{he2017mask}. 
During training and testing, we sample 128 points along each ray and normalize the time indices $i \in \left[0, 1 \right]$. 
As with NeRF~\cite{mildenhall2020nerf}, we use positional encoding to transform the inputs, and parameterize scenes using normalized device coordinates. 
A separate model is trained for a each scene using the Adam optimizer~\cite{Kingma2014AdamAM} with a learning rate of $0.0005$. 
While integrating the static scene representation, we optimize two networks simultaneously. 
Training a full model takes around two days per scene using two NVIDIA V100 GPUs and rendering takes roughly 6 seconds for each $512 \times 288$ frame. 
We refer readers to the supplemental material for our network architectures, hyperparameter settings, and other implementation details.

\begin{table}[tb]
\resizebox{\columnwidth}{!}{%
\begin{tabular}{llcccccc} 
\toprule
\multirow{2}{*}{Methods} & \multirow{2}{*}{MV}
& \multicolumn{3}{c}{Dynamic Only} & \multicolumn{3}{c}{Full} \\
\cmidrule(lr){3-5} \cmidrule(lr){6-8} 
& & SSIM ($\uparrow$) & PSNR ($\uparrow$) & LPIPS ($\downarrow$) & SSIM ($\uparrow$) & PSNR ($\uparrow$) & LPIPS ($\downarrow$) \\ 
\midrule
SinSyn~\cite{Wiles2020SynSinEV} & No & 0.371	& 14.61 &	0.341 & 0.488 &	16.21 &	0.295 \\
MPIs~\cite{single_view_mpi} & No & 0.494 & 16.44	& 0.383 & 0.629 & 19.46	& 0.367 \\
3D Ken Burn~\cite{Niklaus20193DKB}  & No & 0.462 & 16.33	& 0.224 & 0.630 & 19.25 &	0.185 \\
3D Photo~\cite{shih20203d} & No & 0.486 & 16.73 & 0.217 & 0.614 & 19.29	& 0.215 \\
NeRF~\cite{mildenhall2020nerf} & Yes & 0.532 & 16.98 &	0.314 & 0.893 & 24.90	& 0.098 \\
Luo~\etal~\cite{xuan2020consistent} & Yes & 0.530 & 16.97 & 0.207 & 0.746 & 21.37 & 0.141 \\
Yoon~\etal~\cite{yoon2020novel} & Yes & 0.547 & 17.34 & 0.199 & 0.761 & 21.78 & 0.127 \\
\midrule
Ours (wo/ static) & Yes & \textbf{0.760} & \underline{21.88} & \underline{0.108} & \underline{0.906} & \underline{26.95} & \underline{0.071} \\
Ours (w/ static) & Yes & \underline{0.758} & \textbf{21.91} & \textbf{0.097} & \textbf{0.928} & \textbf{28.19} & \textbf{0.045} \\
\bottomrule
\end{tabular}
}
\caption{\textbf{Quantitative evaluation of novel view synthesis on the Dynamic Scenes dataset.} MV indicates whether the approach makes use multi-view information or not.\label{exp:nvidia_quan}}\vspace{-0.2cm}
\end{table}

\begin{table}[tb]
\resizebox{\columnwidth}{!}{%
\begin{tabular}{lccccccccc} 
\toprule
Methods
& \multicolumn{3}{c}{Dynamic Only} & \multicolumn{3}{c}{Full} \\
\cmidrule(lr){2-4} \cmidrule(lr){5-7} 
& SSIM ($\uparrow$) & PSNR ($\uparrow$) & LPIPS ($\downarrow$) & SSIM ($\uparrow$) & PSNR ($\uparrow$) & LPIPS ($\downarrow$) \\ 
\midrule
NeRF~\cite{mildenhall2020nerf}  & 0.522 &	16.74	& 0.328 & 0.862 & 24.29 &	0.113 \\
\cite{shih20203d} + \cite{Niklaus_ICCV_2017_sepconv}  & 0.490 &  16.97 &	0.216 & 0.616 &	19.43 & 0.217 \\
\cite{yoon2020novel} + \cite{Niklaus_ICCV_2017_sepconv} & 0.498 & 16.85 &	0.201 & 0.748	& 21.55	& 0.134 \\
\midrule
Ours (w/o static) & \underline{0.720} & \underline{21.51} & \underline{0.149} & \underline{0.875} & \underline{26.35} & \underline{0.090} \\
Ours (w/ static) & \textbf{0.724} & \textbf{21.58} & \textbf{0.143} & \textbf{0.892} & 	\textbf{27.38} & \textbf{0.066}  \\
\bottomrule
\end{tabular}
}
\caption{\textbf{Quantitative evaluation of novel view and time synthesis.} See Sec.~\ref{sec:quant_eval} for a description of the baselines.}\vspace{-0.2cm}
\label{exp:nvidia_quan_space_time}
\end{table}

\subsection{Baselines and error metrics}
We compare our approach to state-of-the-art single-view and multi-view novel view synthesis algorithms.
For single-view methods, we compare to MPIs~\cite{single_view_mpi} and SinSyn~\cite{Wiles2020SynSinEV}, trained on indoor real estate videos~\cite{zhou2018stereo}; 3D Photos~\cite{shih20203d} and 3D Ken Burns~\cite{Niklaus20193DKB} were trained mainly on images in the wild.
Since these methods can only compute depth up to an unknown scale and shift, we align the predicted depths with the SfM sparse point clouds before rendering.
For multi-view, we compare to a recent dynamic view synthesis method~\cite{yoon2020novel}.
Since the authors do not provide source code, we reimplemented their approach based on the paper description.
We also compare to a video depth prediction method~\cite{xuan2020consistent} and perform novel view synthesis by rendering the point cloud into novel views while filling in disoccluded regions.
Finally, we train a standard NeRF~\cite{mildenhall2020nerf}, with and without the added time domain, on each dynamic scene.

We report the rendering quality of each approach with three standard error metrics: structural similarity index measure (SSIM),  peak signal-to-noise ratio (PSNR), and perceptual similarity through LPIPS~\cite{zhang2018unreasonable}, on both the entire scene (Full) and in dynamic regions only (Dynamic Only). 

\subsection{Quantitative evaluation}
\label{sec:quant_eval}
We evaluate on the Nvidia Dynamic Scenes Dataset~\cite{yoon2020novel}, which consists of 8 scenes with human and non-human motion recorded by 12 synchronized cameras.
As in the original work~\cite{yoon2020novel}, we simulate a moving monocular camera by extracting images sampled from each camera viewpoint at different time instances, and evaluate the result of view synthesis with respect to known held-out viewpoints and frames. For each scene, we extract 24 frames from the original videos for training and use the remaining 11 held-out images per time instance for evaluation.

\paragraph{Novel view synthesis.}
We first evaluate our approach and other baselines on the task of novel view synthesis (at the same time instances as the training sequences).
The quantitative results are shown in Table~\ref{exp:nvidia_quan}.
Our approach without the static scene representation (Ours w/o static) already significantly outperforms other single-view and multi-view baselines in both dynamic regions and on the entire scene. 
NeRF has the second best performance on the entire scene, but cannot model scene dynamics.
Moreover, adding the static scene representation improves overall rendering quality by more than 30\%, demonstrating the benefits of leveraging global multi-view information from rigid regions where possible. 

\paragraph{Novel view and time synthesis.}
We also evaluate the task of novel view and time synthesis by extracting every other frame from the original Dynamic Scenes dataset videos for training, and evaluating on the held-out intermediate time instances at held-out camera viewpoints.
Since we are not aware of prior monocular space-time view interpolation methods, we use two state-of-the-art view synthesis baselines~\cite{shih20203d, yoon2020novel} to synthesize images at the testing camera viewpoints followed by 2D frame interpolation~\cite{Niklaus_ICCV_2017_sepconv} to render intermediate times, as well as NeRF evaluated directly at the novel space-time views.
Table~\ref{exp:nvidia_quan_space_time} shows that our method significantly outperforms all baselines in both dynamic regions and the entire scene.

\paragraph{Ablation study.}  \label{sec:ablation}
We analyze the effect of each proposed system component in the task of novel view synthesis by removing (1) all added losses, which gives us NeRF extended to the temporal domain (NeRF (w/ time)); (2) the single view depth prior (w/o $\Ldisp$); (3) the geometry consistency prior (w/o $\Lgeo$); (4) the scene flow cycle consistency term (w/o $\Lcycle$); (5) the scene flow regularization term (w/o $\Lphy$); (6) the disocculusion weight fields (w/o $\weightSet$); (7) the static representation (w/o static).
The results, shown in Table~\ref{exp:ablation_nvidia}, demonstrate the relative importance of each component, with the full system performing the best.

\begin{table}[tb]
\resizebox{\columnwidth}{!}{%
\begin{tabular}{lccccccccc} 
\toprule
Methods
& \multicolumn{3}{c}{Dynamic Only} & \multicolumn{3}{c}{Full} \\
\cmidrule(lr){2-4} \cmidrule(lr){5-7} 
& SSIM ($\uparrow$) & PSNR ($\uparrow$) & LPIPS ($\downarrow$) & SSIM ($\uparrow$) & PSNR ($\uparrow$) & LPIPS ($\downarrow$) \\ 
\midrule
NeRF (w/ \text{time}) & 0.630 & 18.89	& 0.159 & 0.875 & 24.33 & 0.081 \\
w/o $\Ldisp$ & 0.710 & 19.66	& 0.132 & 0.882 & 25.16 &	0.078 \\
w/o $\Lgeo$  & 	0.713	& 19.74 &	0.139 & 0.885 & 25.19 & 0.079 \\
w/o $\Lcycle$   & 0.731	& 20.52 & 0.115 & 0.890	& 26.15 & 0.072 \\
w/o $\Lphy$   & 0.751 &	21.22	& 0.110 & 0.895	& 26.67 & 0.074   \\
w/o $\weightSet$  & 0.754 & 21.31	& 0.112 & 0.894	& 26.23	& 0.074 \\
w/o static & \textbf{0.760} & \underline{21.88} & \underline{0.108} & \underline{0.906} & \underline{26.95} & \underline{0.071} \\
Full (w/ static) & \underline{0.758} & \textbf{21.91} & \textbf{0.097} & \textbf{0.928} & \textbf{28.19} & \textbf{0.045}  \\
\bottomrule
\end{tabular}
}
\caption{\textbf{Ablation study on the Dynamic Scenes dataset.} See Sec.~\ref{sec:ablation} for detailed descriptions of each of the ablations.}\vspace{-0.2cm}
\label{exp:ablation_nvidia}
\end{table}

\subsection{Qualitative evaluation}

\begin{figure*}[t]
    \centering
    \setlength{\tabcolsep}{0.025cm}
    \setlength{\itemheight}{2.55cm}
    \renewcommand{\arraystretch}{0.5}
    \hspace*{-\tabcolsep}\begin{tabular}{cccccc}
            \includegraphics[height=\itemheight]{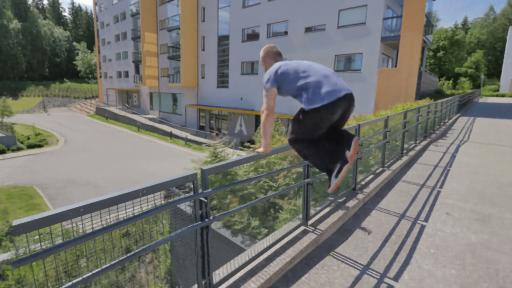}
        &
            \includegraphics[height=\itemheight]{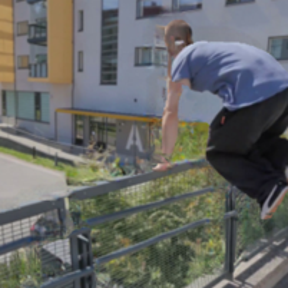}
        &
            \includegraphics[height=\itemheight]{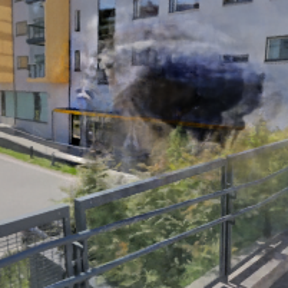}
        &
            \includegraphics[height=\itemheight]{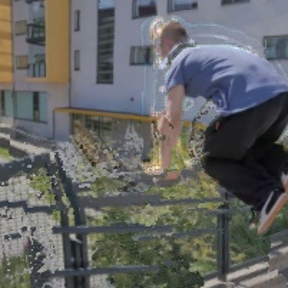}
        &
            \includegraphics[height=\itemheight]{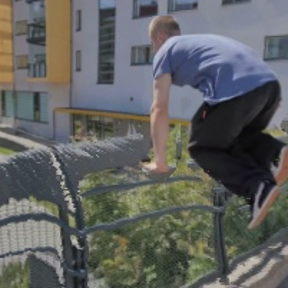}
        &
            \includegraphics[height=\itemheight]{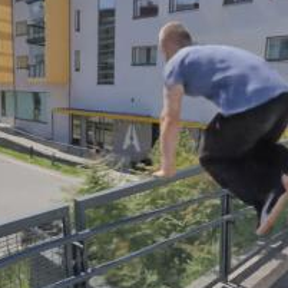}
        \\
            \includegraphics[height=\itemheight]{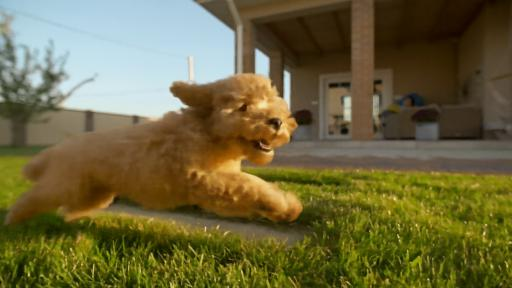}
        &
            \includegraphics[height=\itemheight]{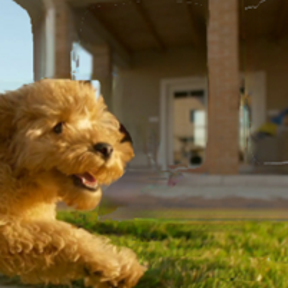}
        &
            \includegraphics[height=\itemheight]{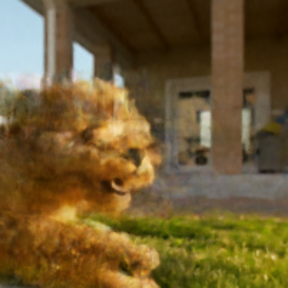}
        &
            \includegraphics[height=\itemheight]{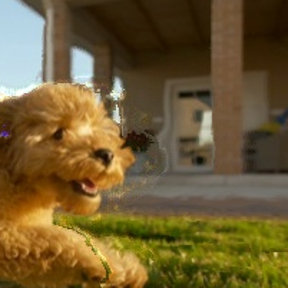}
        &
            \includegraphics[height=\itemheight]{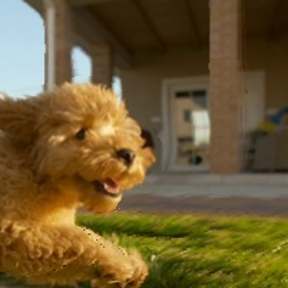}
        &
            \includegraphics[height=\itemheight]{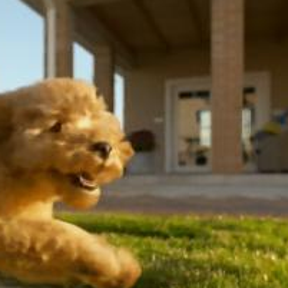}
        \\
            \includegraphics[height=\itemheight]{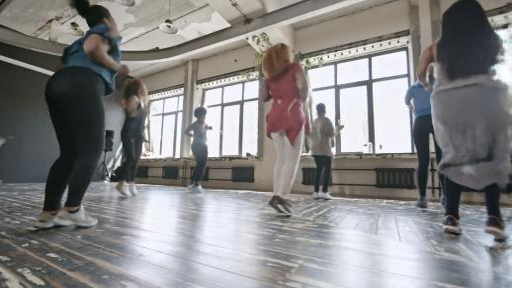}
        &
            \includegraphics[height=\itemheight]{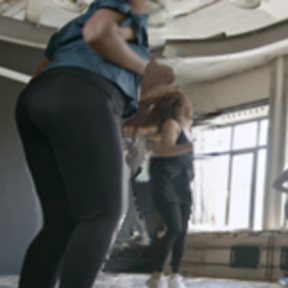}
        &
            \includegraphics[height=\itemheight]{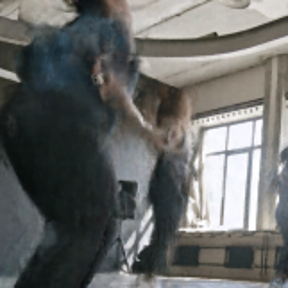}
        &
            \includegraphics[height=\itemheight]{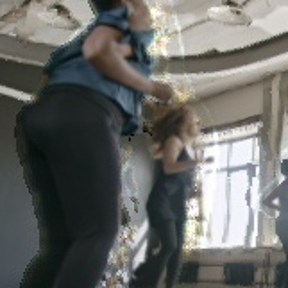}
        &
            \includegraphics[height=\itemheight]{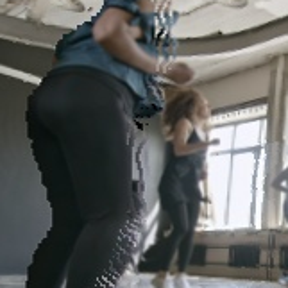}
        &
            \includegraphics[height=\itemheight]{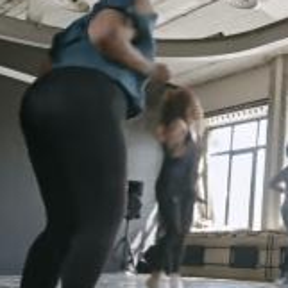}
        \\
            \includegraphics[height=\itemheight]{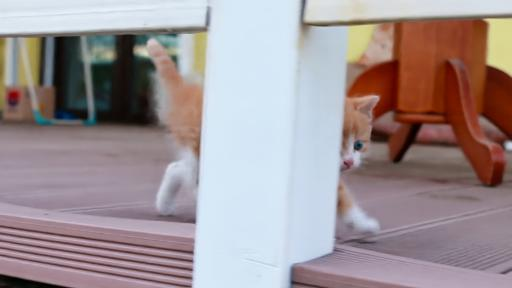}
        &
            \includegraphics[height=\itemheight]{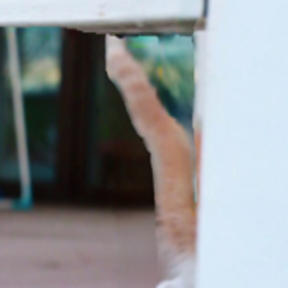}
        &
            \includegraphics[height=\itemheight]{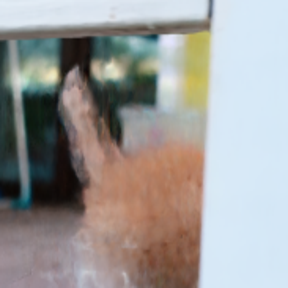}
        &
            \includegraphics[height=\itemheight]{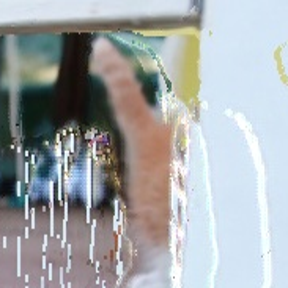}
        &
            \includegraphics[height=\itemheight]{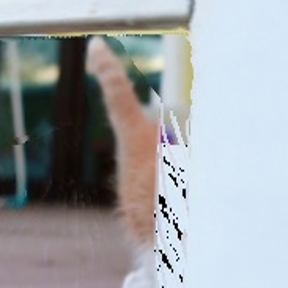}
        &
            \includegraphics[height=\itemheight]{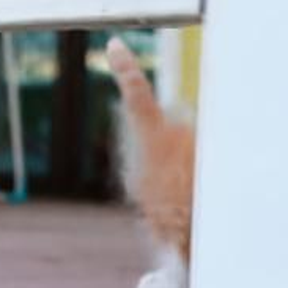}
        \\
            \vphantom{\large I} Our rendered views
        &
            \vphantom{\large I} 3D Photo~\cite{shih20203d}
        &
            \vphantom{\large I} NeRF~\cite{mildenhall2020nerf}
        &
            \vphantom{\large I} Yoon~\etal~\cite{yoon2020novel}
        &
            \vphantom{\large I} Luo~\etal~\cite{xuan2020consistent}
        &
            \vphantom{\large I} Ours
        \\
    \end{tabular}\vspace{-0.2cm}
    \caption{\textbf{Qualitative comparisons on monocular video clips.} When compared to baselines, our approach more correctly synthesizes hidden content in disocclusions (shown in the last three rows), and locations with complex scene structure such as the fence in the first row.  \label{fig:monocular_qual}}\vspace{-0.2cm}
\end{figure*}

We provide qualitative comparisons on the Dynamic Scenes dataset (Fig.~\ref{fig:nvidia_qual}) and on monocular video clips collected in-the-wild from the internet featuring complex object motions such as jumping, running, or dancing with various occlusions (Fig.~\ref{fig:monocular_qual}).
NeRF~\cite{mildenhall2020nerf} correctly reconstructs most static regions, but produces ghosting in dynamic regions since it treats all the moving objects as view-dependent effects, leading to incorrect interpolation results.
The state-of-the-art single-view method~\cite{shih20203d} tends to synthesize incorrect content at disocclusions, such as the bins and speaker in the last three rows of Fig.~\ref{fig:monocular_qual}. 
In contrast, methods based on reconstructing explicit depth maps~\cite{xuan2020consistent, yoon2020novel} have difficulty modeling complex scene appearance and geometry such as the thin structures in the third row of Fig.~\ref{fig:nvidia_qual} and the first row of Fig.~\ref{fig:monocular_qual}.

\section{Discussion}

\begin{figure}[t]
    \centering
    \setlength{\tabcolsep}{0.025cm}
    \setlength{\itemwidth}{2.73cm}
    \renewcommand{\arraystretch}{0.5}
    \hspace*{-\tabcolsep}\begin{tabular}{ccc}
            \begin{tikzpicture}
                \definecolor{boxcolor}{RGB}{255,0,0}
                \node [anchor=south west, inner sep=0.0cm] (image) at (0,0) {
                    \includegraphics[width=\itemwidth]{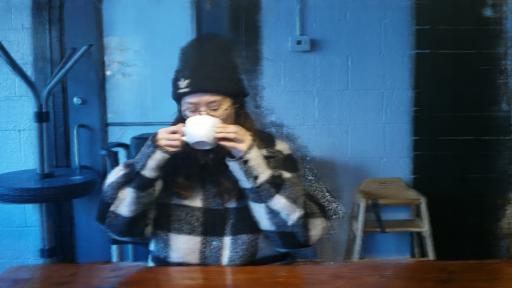}
                };
                \begin{scope}[x={(image.south east)},y={(image.north west)}]
                    \draw [boxcolor] (0.5,0.60) rectangle (0.7,0.18);
                \end{scope}
            \end{tikzpicture}
        &
            \begin{tikzpicture}
                \definecolor{boxcolor}{RGB}{255,0,0}
                \node [anchor=south west, inner sep=0.0cm] (image) at (0,0) {
                    \includegraphics[width=\itemwidth]{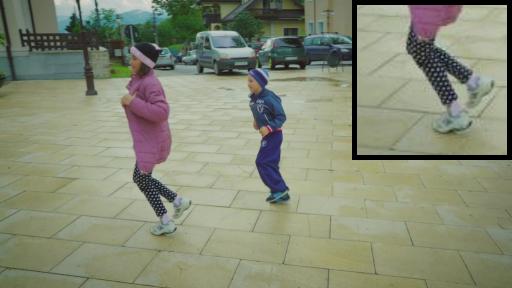}
                };
                \begin{scope}[x={(image.south east)},y={(image.north west)}]
                    \draw [boxcolor] (0.2,0.5) rectangle (0.4,0.15);
                \end{scope}
            \end{tikzpicture}
        &
            \includegraphics[width=\itemwidth]{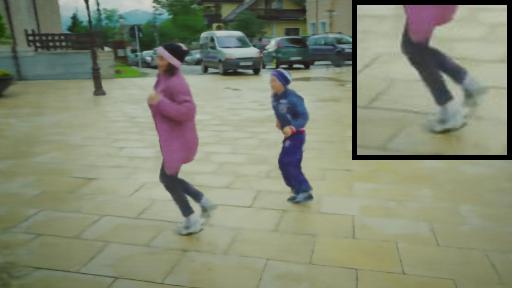}
        \\
            \vphantom{\small I}\scriptsize (a) Non-seen disocclusion
        &
            \vphantom{\small I}\scriptsize (b) GT for (c)
        &
            \vphantom{\small I}\scriptsize (c) Missing details
        \\
    \end{tabular} \vspace{-0.25cm}
    \caption{\textbf{Limitations.} Our method is unable to extrapolate content unseen in the training views (a), and has difficulty recovering high frequency details if a video involves extreme object motions (b,c).} \vspace{-0.2cm}
    \label{fig:limitations}
\end{figure}

\paragraph{Limitations.}
Monocular space-time view synthesis of dynamic scenes is  challenging, and we have only scratched the surface with our proposed method. 
In particular, there are several limitations to our approach.
Similar to NeRF, training and rendering times are high, even at limited resolutions. Additionally, each scene has to be reconstructed from scratch and our representation is unable to extrapolate content unseen in the training views (See Fig.~\ref{fig:limitations}(a)).
Furthermore, we found that rendering quality degrades when either the length of the sequence is increased given default number of model parameters (most of our sequences were trained for 1$\sim$2 seconds), or when the amount of motion is extreme (See Fig.~\ref{fig:limitations}(b-c), where we train a model on a low frame rate video).
Our method can end up in the incorrect local minima if object and camera motions are close to a degenerate case, e.g., colinear, as described in Park~\etal~\cite{park20103d}.

\paragraph{Conclusion.}
We presented an approach for monocular novel view and time synthesis of complex dynamic scenes by Neural Scene Flow Fields, a new representation that implicitly models scene time-variant reflectance, geometry and 3D motion.
We have shown that our method can generate compelling space-time view synthesis results for scenes with natural in-the-wild scene motion. 
In the future, we hope that such methods can enable high-resolution views of dynamic scenes with larger scale and larger viewpoint changes.

{\small
\bibliographystyle{ieee_fullname}
\bibliography{dynerf}
}

\end{document}